\documentclass[letterpaper]{article} 
\usepackage{aaai2026}  
\usepackage{times}  
\usepackage{helvet}  
\usepackage{courier}  
\usepackage[hyphens]{url}  
\usepackage{graphicx} 
\usepackage{natbib}  
\usepackage{caption} 
\usepackage{algorithm}
\usepackage{algorithmic}
\usepackage{amsthm}
\usepackage{amsmath}
\usepackage{subfigure}
\usepackage{array}
\usepackage{multirow}
\usepackage{amsfonts}
\usepackage{booktabs} 
\usepackage{amssymb}
\usepackage{enumitem}
\newtheorem{theorem}{Theorem}
\newtheorem{definition}[theorem]{Definition} 
\newtheorem{lemma}[theorem]{Lemma} 
 
\newtheorem{corollary}[theorem]{Corollary}
 \newcolumntype{Y}{>{\centering\arraybackslash}X} 
 \usepackage{tabularx}
\usepackage{newfloat}
\usepackage{listings}
\DeclareCaptionStyle{ruled}{labelfont=normalfont,labelsep=colon,strut=off} 
\floatstyle{ruled}
\newfloat{listing}{tb}{lst}{}
\floatname{listing}{Listing}
\newenvironment{manualtheorem}[2]{%
  \par\addvspace{1em}%
  \noindent\textbf{#1~#2.} \itshape%
}{\par\addvspace{1em}}

 \newcolumntype{Y}{>{\centering\arraybackslash}X} 
 \usepackage{tabularx}
\usepackage{newfloat}
\usepackage{listings}
\DeclareCaptionStyle{ruled}{labelfont=normalfont,labelsep=colon,strut=off} 
\floatstyle{ruled}
\newfloat{listing}{tb}{lst}{}
\floatname{listing}{Listing}

\title{Enhancing Logical Expressiveness in Graph Neural Networks\\ via
Path-Neighbor Aggregation}
\author{
    Han Yu\textsuperscript{\rm 1},
    Xiaojuan Zhao\textsuperscript{\rm 2 *},
    Aiping Li\textsuperscript{\rm 1}\thanks{\ Corresponding authors.},Kai Chen\textsuperscript{\rm 1},Ziniu Liu\textsuperscript{\rm 1}, Zhichao Peng\textsuperscript{\rm 2}
}

\affiliations{
    \textsuperscript{\rm 1}College of Computer Science and Technology, National University of Defense Technology, Changsha, China.\\

    \textsuperscript{\rm 2}Information School, Hunan University of Humanities, Science and Technology, Loudi, China.\\

     \{yuhan17,zhaoxiaojuan18,liaiping,chenkai\_, liuzn\_nudt\}@nudt.edu.cn, zcpeng@tju.edu.cn
}

\usepackage{bibentry}

\begin{document}

\maketitle

\begin{abstract}
Graph neural networks (GNNs) can effectively model structural information of graphs, making them widely used in knowledge graph (KG) reasoning. However, existing studies on the expressive power of GNNs mainly focuses on simple single-relation graphs, and there is still insufficient discussion on the power of GNN to express logical rules in KGs. How to enhance the logical expressive power of GNNs is still a key issue. Motivated by this, we propose Path-Neighbor enhanced GNN (PN-GNN), a method to enhance the logical expressive power of GNN by aggregating node-neighbor embeddings on the reasoning path. First, we analyze the logical expressive power of existing GNN-based methods and point out the shortcomings of the expressive power of these methods. 
Then, we theoretically investigate the logical expressive power of PN-GNN, showing that it not only has strictly stronger expressive power than C-GNN but also that its $(k+1)$-hop logical expressiveness is strictly superior to that of $k$-hop. Finally, we evaluate the logical expressive power of PN-GNN on eight synthetic datasets and two real-world datasets. Both theoretical analysis and extensive experiments confirm that PN-GNN enhances the expressive power of logical rules without compromising generalization, as evidenced by its competitive performance in KG reasoning tasks.
\end{abstract}

\section{Introduction}

A Knowledge Graph (KG) organizes and represents knowledge in a structured form. In a KG, the nodes represent entities (such as people, places, or concepts), while the edges capture relationships between entities (such as ``belongs to'' or ``contains''). 
KG reasoning typically requires models to understand and infer complex relations within graph structures. In recent years, graph neural networks (GNNs) \cite{wu2020comprehensive,velivckovic2017graph,hamilton2017inductive,xu2018powerful} have shown great potential in KG reasoning tasks due to their powerful ability to represent graph-structured data. GNNs can automatically learn intricate relations between nodes from data and effectively capture both global and local structural information via message passing mechanisms, thereby substantially enhancing reasoning capabilities.

With the widespread application of GNNs in KG reasoning, their expressive power is crucial and can be evaluated from two aspects. One is discrimination, that is, the ability to distinguish non-isomorphic graphs, which is closely related to the strength of the Weisfeiler-Leman test. Another is to directly characterize what specific function classes a model can approximate, that is, the ability to learn specific logical rule structures. In KG reasoning tasks, if GNN cannot learn certain key rule structures, its performance may be severely limited. 
\begin{figure}[h]
  \centering
\includegraphics[width=0.98\linewidth]{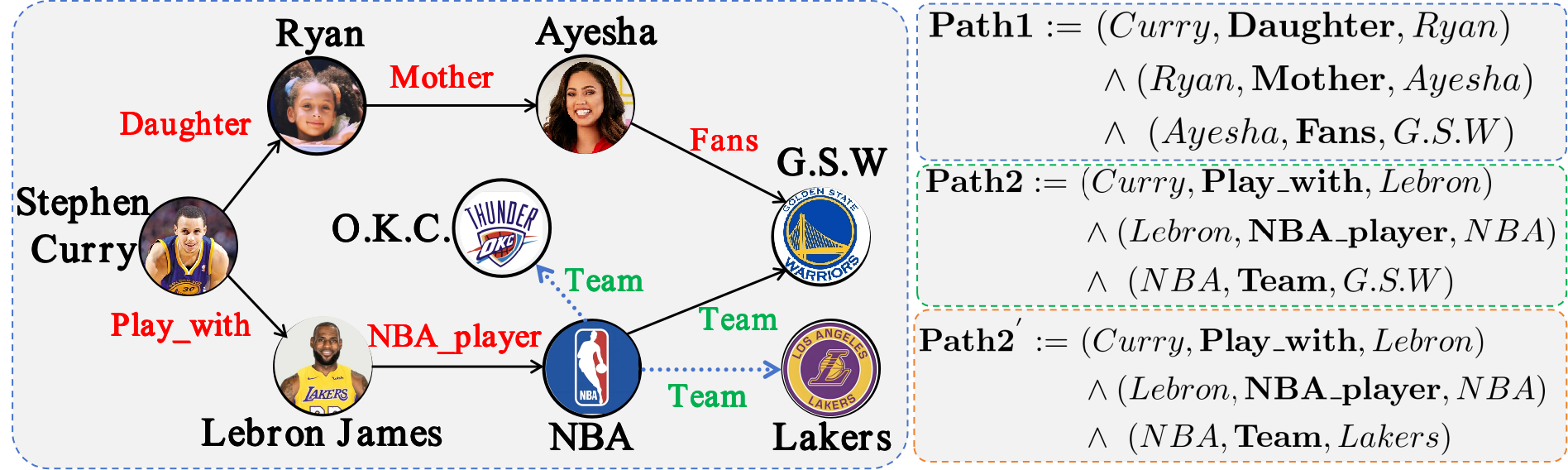}
  \caption{Schematic diagram of knowledge graph reasoning.  
  }
  \label{fig:intro1}
\end{figure}
As shown in Figure \ref{fig:intro1}, to reason the triple $(Curry, Team, ?)$. Relying solely on either Path1 or Path2 is not sufficient to correctly infer the answer. 
Path1 only shows that Curry’s family members are fans of G.S.W., but does not indicate whether Curry is an NBA player or which team he belongs to. 
Path2 shows that Curry is an NBA player, but not which specific team he plays for. 
Path2 and Path2' share the same relation path ($Play\_with$$\rightarrow$$NBA\_player$$\rightarrow$$Team$), and Path2' incorrectly links Curry to the Lakers. 
Only when both Path1 and Path2 are simultaneously satisfied can we infer the conclusion of ($Curry$, $Team$, $G.S.W$).
Therefore, if the model's expressive power is insufficient to capture and reason about these relations, it may fail to make correct predictions in reasoning tasks.
This not only limits the model's reasoning ability but also leads to poor performance in real scenarios. The limited expressive power of key rule structures significantly affects the model’s effectiveness and applicability. Thus, this motivates developing more powerful GNN models by enhancing the expressiveness of logical rules.

From the perspective of the logical expression ability of GNN, relational aggregation methods such as R-GCN \cite{schlichtkrull2018modeling} and CompGCN \cite{vashishth2019composition}, cannot perform rule learning well. The GNN method that dynamically adjusts based on conditions, such as NBFNet \cite{zhu2021neural}, RED-GNN \cite{zhang2022knowledge}, and A*Net \cite{zhu2024net}, has been proven to have strong logical expressiveness \cite{understanding, ATheory} and can learn basic logical relationships, such as unary prediction, negation, conjunction, disjunction, etc.
Based on the Relational Weisfeiler-Leman (R-WL) \cite{Barcelo00O22} test, 
the limitations of the logical rule learning ability of existing relational GNN (R-GNN) models were clarified, and it was proved that these models cannot capture higher-order logical rules \cite{ATheory}. In other words, it was theoretically proved that the rule expression ability of conditional GNN (C-GNN) is strictly more powerful than that of R-GNN method. 
Although the labeling trick can enhance the rule learning ability of C-GNN in KG link prediction by introducing specific labels to nodes, it still faces two limitations: on the one hand, it will reduce the generalization ability of the model; on the other hand, the design of adding constant labels makes it difficult to apply to KG reasoning tasks in inductive settings. 
\begin{figure}[h]
  \centering
  \includegraphics[width=0.6\linewidth]{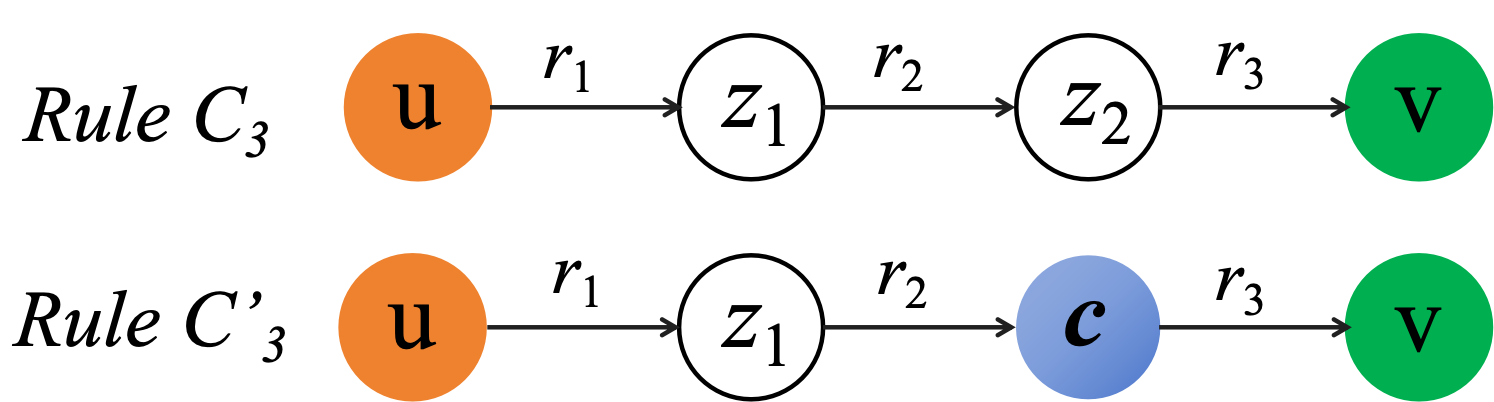}
  \caption{Rules $C_3$ and $C'_3$. 
  Here, $z_1$, $ z_2 $ are variables and $c$ is a constant.
  }
  
  \label{fig:intro}
\end{figure}
For example, in Figure \ref{fig:intro}, for the rule 
$C_{3}(u,v):= \exists\,z_{1},z_{2} (
  r_{1}(u,z_{1})  \wedge  r_{2}(z_{1},z_{2})  \wedge  r_{3}(z_{2},v)
)$,
when $z_{1}$ and $z_{2}$ in the rule are variables, it can match any element that satisfies the variable constraint; but once the variable is fixed to a constant \textit{c}, that is, it becomes $C_{3}'(u,v)  :=  \exists\ z_{1} ( r_{1}(u,z_{1})  \wedge  r_{2}(z_{1},\textit{c}) \wedge r_{3}(\textit{c},v))$, only the specific elements corresponding to \textit{c }can satisfy the rule, so the scope of application of the rule is limited, and the coverage \cite{Quinlan1990, MUGGLETON1994629} naturally becomes smaller, thereby weakening the generalization ability of the model.

In this paper, we propose \textbf{P}ath-\textbf{N}eighbor enhanced \textbf{GNN} (\textbf{PN-GNN}), a method to enhance the logical expressive power of GNN. PN-GNN addresses the labeling trick’s limitations while preserving the model’s generalization. 
In the conditional message passing scenario, PN-GNN leverages neighboring nodes along the paths between the source node and each target node to enrich the model’s structural information, thereby enhancing the logical expressiveness. PN-GNN not only improves logical reasoning capabilities but also preserves generalization ability, making it suitable for inductive settings. In summary, our key contributions are:
\begin{itemize} 
\item We propose a method called Path-Neighbor Enhanced GNN (PN-GNN) that significantly improves the logical expressive power without compromising the generalization ability, and PN-GNN can be effectively applied in inductive settings.
\item  We theoretically analyze the logical expressive power of PN-GNN and demonstrate that it surpasses that of GNNs based on conditional message passing.
\item We instantiate the proposed theoretical model PN-GNN and build a KG reasoning model based on PN-GNN. The effectiveness of PN-GNN is verified by constructing synthetic datasets and real-world datasets. It outperforms the benchmark dataset in all indicators.
\end{itemize}

\section{Related Work}

\subsection{Expressivity of Graph Neural Networks}

The expressive power of GNNs is often analyzed through graph isomorphism testing. Vanilla GNNs have been shown to match the 1-WL test \cite{xu2018powerful}, with extensions to KGs \cite{Barcelo00O22}. To enhance expressivity, two main directions have emerged: designing stronger WL variants \cite{morris2019weisfeiler,otto2023graded,Barcelo00O22}, and applying initialization techniques such as node labeling or identity features 
\cite{AbboudCGL21,you2021identity,SatoYK21}.
In link prediction, the structural expressivity of GNN alone has been found to be insufficient \cite{Srinivasan020}, and the incorporation of labeling enables GNNs to learn structural information, although the existing results are primarily limited to single relational graphs \cite{LabelingTrick}.
Beyond WL-based approaches, GNNs have been shown to model first-order logic (FOL) with two variables ($\text{FOC}_2$), and their expressivity in node classification can be described using the counting extension of modal logic \cite{barcelo2020logical,cai1992optimal}. This has been extended to KGs, linking logical and WL-based expressivity and identifying the rule structures that GNNs can learn \cite{ATheory}. 
Inspired by Teru, Denis, and Hamilton (2020), further improvements combine conditional message passing with labeling to enhance the learning of logical rules in KG reasoning.

\subsection{Knowledge Graph Reasoning based on GNN}
\label{sec:4}
GNN-based KG reasoning can be categorized into R-GNN and C-GNN \cite{ATheory}. R-GNN methods like R-GCN and CompGCN perform relation-aware message passing to update entity embeddings, which are then used to score candidate links between entities.

C-GNN methods generate query-aware representations by marking the head entity and propagating messages accordingly, and can be categorized into subgraph-based and progressive aggregation methods.
Subgraph-based models, such as GraIL \cite{teru2020inductive} and CoMPILE \cite{mai2021communicative}, extract local neighborhoods between entity pairs and apply GNNs to propagate messages within subgraphs.
Progressive aggregation methods, such as NBFNet \cite{zhu2021neural}, RED-GNN \cite{zhang2022knowledge}, and A*Net \cite{zhu2024net}, further enhance reasoning capabilities. NBFNet incorporates the Bellman-Ford \cite{goldberg1993heuristic} to propagate messages along multi-hop paths. 
RED-GNN determines the next hop based on the current node, the query relation, and the next relation embedding.
A*Net introduces A* search to enhance inference efficiently.

\section{Background}

\subsection{Knowledge graphs and Binary Predicate}

A knowledge graph $\mathcal{G} = \{\mathcal{E},\mathcal{R},\mathcal{T} \}$, where $\mathcal{E}$ represents the set of entities (nodes), $\mathcal{R}$ represents the set of relations (edges). $\mathcal{T} \subseteq \mathcal{E}\times \mathcal{R} \times \mathcal{E}$ represents set of triplets and each triplet can be represented as $(s,r,o)$, $ s,o \in \mathcal{E}$, $ r \in \mathcal{R}$. 
A constant ($ \textit{c} \in \mathcal{E}$) is a unique identifier for an entity in KG, while a variable ($ z_{i} \in \mathcal{E}$) represents any entity in KG. The binary predicate $ r_j(s,o) $, where $r_j$ ($r_j \in \mathcal{R}$) represents a certain relation or attribute.

\subsection{Conditional Graph Neural Networks}

Graph neural networks based on conditional message passing mechanisms \cite{zhu2021neural, zhang2022knowledge} 
have a common message passing mechanism. 
Given a triple query $(u, q, ?$), where \textit{u} is the head entity and \textit{q} is the query relation. Conditional Graph Neural Networks (C-GNN) first mark the head entity \textit{u} with the query relation \textit{q} to distinguish it from all other nodes in the KG. 
Then, it iteratively calculates and updates the representations of entity pairs ${\textbf{h}}_{v|u,q}$ of all entities in the KG that can reach the tail \textit{v} from the head \textit{u}:
\begin{equation}
\begin{aligned}
	{\textbf{h}}_{v|u,q}^{(0)} = &\text{INIT}(u,v,q)\\
	{\textbf{h}}_{v|u,q}^{(t+1)}=&
    \underset{\substack{w \in N_r(v) \\ r \in R}}
    {\text{UPDATE}}
    \bigg(
    {\textbf{h}}_{v|u,q}^{(t)},
    \text{AGG}\Big(
    \{\{\text{MSG}_{r}({\textbf{h}}_{w|u,q}^{(t)},{\textbf{z}}_{\textbf{q}})\}\}\Big)
    \bigg)
\end{aligned}
\label{nbfnet}
\end{equation}

Here, INIT denotes the initialization function, UPDATE represents the update function, $N_{r}(v)$ denotes the set of neighbors connected to $v$ via relation $r$, $\textbf{z}_\textbf{q}$ denotes the embedding of the query relation. AGG is the aggregation function, while MSG refers to the message passing function. The notation $\{\{\cdot\}\}$ represents a multi-set. C-GNN obtains the pairwise-entity representations ${\textbf{h}}_{v|u,q}^{(L)}$ of the tail entity $v$, conditioned on the head entity $u$ and the query relation $q$, through an \textit{L}-layer message passing process. Finally, C-GNN predicts the target based on the tail representation:
\begin{equation}
 score(u,q,v) = \sigma(\text{MLP}({\textbf{h}}_{v|u,q}^{(L)})) 
\end{equation}
where $\text{MLP}(\cdot)$ represents a trainable multi-layer perceptron.

\section{Logical Expressive Power of Conditional GNN}
\label{sec:LEP}
The logical expressivity of a GNN refers to the set of logical formulas it can learn \cite{understanding}. C-GNN demonstrates superior logical expressive power compared to R-GNN, leading to enhanced performance \cite{ATheory,understanding}.  
We use the counting extension of modal logic (CML), which extends the counting expressiveness of graded modal logic, to describe FOL properties in KG.
CML consists of propositional constants (always true $\top$, always false $\bot$), unary logical predicates $P_{i}(x)$, and recursive rules. Given FOL $\varphi(x,y)$, $\psi(x,y)$, positive integer variables $N \geq 1$ and $r \in R$, according to the recursive rules, $\neg \varphi(x,y)$, $\varphi(x,y) \wedge \psi(x,y)$, and $\exists^{N \geq n}z(\varphi(x,z) \wedge r(z,y))$ are also in CML.
Based on the CML in KG, the logical rule expressive power of C-GNN can be expressed as follows:
\begin{theorem}
	\label{theorem:CGNN1}
C-GNN can learn every CML-based formula $\varphi(x)$ by leveraging its entity representations.
\end{theorem}

\begin{theorem}
	\label{theorem:CGNN2}
A formula $\varphi(x)$ is learned by C-GNN if it can be expressed as a formula in CML.
\end{theorem}

Theorem \ref{theorem:CGNN1} shows that C-GNN has the ability to learn all CML formulas, implying that its logical expressive power of C-GNN at least covers the set of CML formulas. Conversely, Theorem \ref{theorem:CGNN2} restricts the expressive power of C-GNN, by showing that it can only learn formulas that can be converted into CML formulas. Together, these two theorems characterize the logical expressive power of C-GNN, which is equivalent to that of CML formulas. See Appendix A.1 for the complete proofs.

\begin{figure}[t]
    \centering
    \subfigure[$C_{3}(u,v)$]{
        \includegraphics[width=0.2\textwidth]{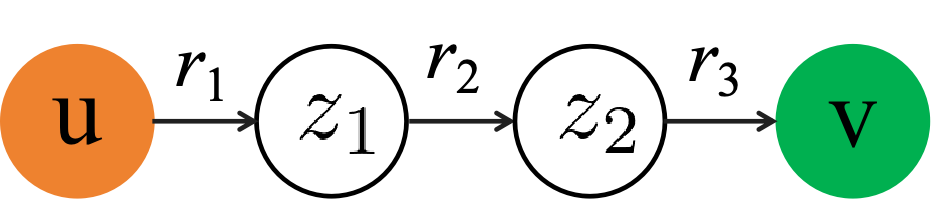}
        \label{fig:a}
    }
    \subfigure[$I_{2}(u,v)$]{
        \includegraphics[width=0.20\textwidth]{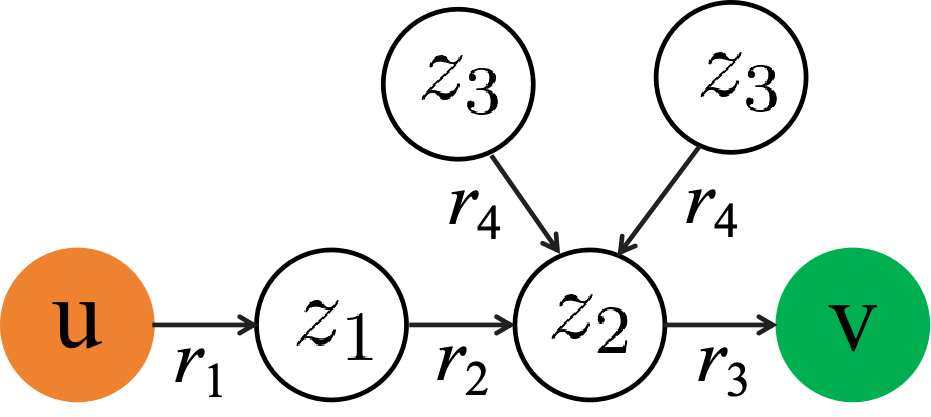}
        \label{fig:b}
    }
    \subfigure[$T(u,v)$]{
        \includegraphics[width=0.20\textwidth]{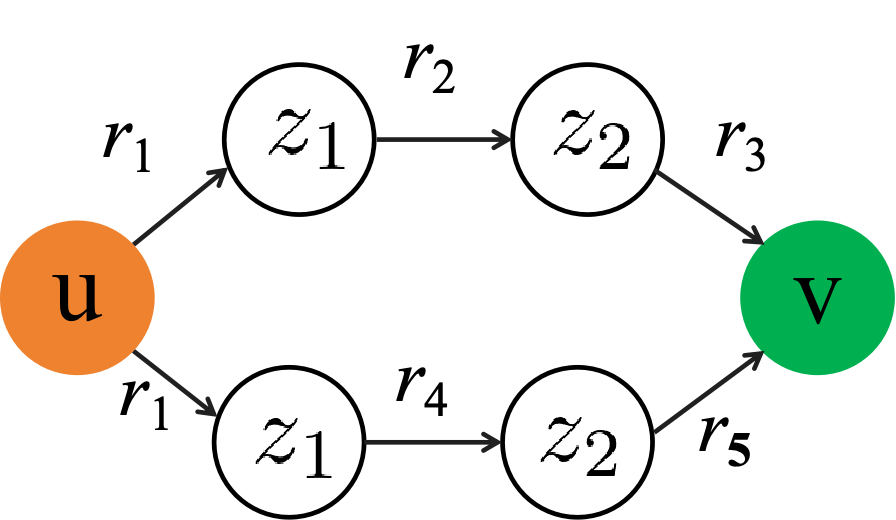}
        \label{fig:c}
    }
    \subfigure[$U(u,v)$]{
        \includegraphics[width=0.20\textwidth]{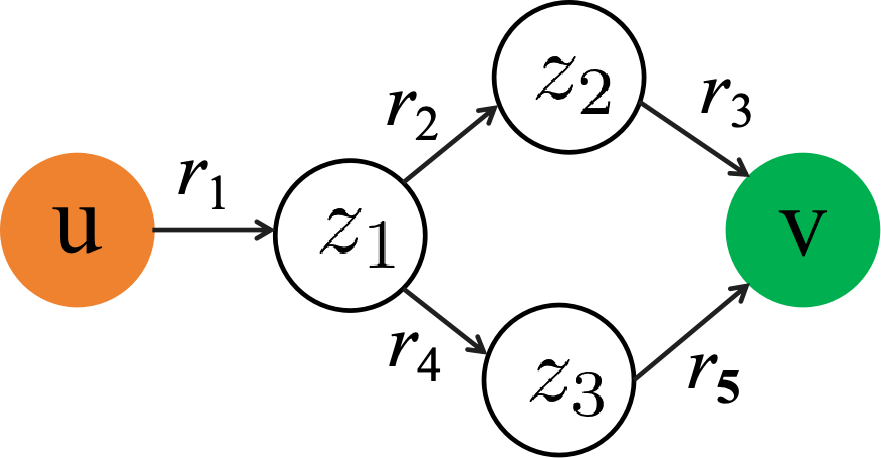}
        \label{fig:d}
    }
    \caption{The rule structure that C-GNN can learn, taking 3-hop as an example. }
    \label{fig:all}
\end{figure}

In KGs, CML can represent several common logical rule structures, as shown in Figure \ref{fig:all}. 
Chain-like rules $ C_3$, as illustrated in Figure \ref{fig:a}, represent a commonly used rule structure in KG reasoning methods that rely on paths or rules \cite{understanding,ATheory}. The corresponding rule formula is expressed as follows:
\begin{equation}
C_{3}(u,v)  := \exists\,z_{1},z_{2}(
  r_{1}(u,z_{1}) \wedge  r_{2}(z_{1},z_{2})  \wedge  r_{3}(z_{2},v)
).
\end{equation}
Since $ C_3$ can be described by CML based on recursion, C-GNN can learn the chain logic rule structure $ C_3$. 

The second rule structure $I_{N}$ adds one or more relations $r \in \mathcal{R} $ that need to be satisfied to the chain logic rule, as shown in Figure \ref{fig:b}, the case for $N=2$ is illustrated and denoted as $I_2$. In the KG, $I_{N}$ represents the scenario in which the rule is valid only if specific conditions are satisfied on certain entities. 
$I_{N}$ can be described by CML, so it can be learned by C-GNN.
\begin{multline}
I_{N}(u,v)  := 
  \exists\,z_{1},z_{2} (r_{1}(u,z_{1})  \wedge  r_{2}(z_{1},z_{2}))\\
   \wedge \{\exists^{\ge N} z_{3}( r_{4}(z_{3},z_{2}))\}
   \wedge r_{3}(z_{2},v).
\end{multline}

The third rule structure $T$, as shown in Figure \ref{fig:c}, the upper and lower parts of the structure can be regarded as chain rules respectively, which can be described by CML. According to the recursive rule of CML, we can merge these two parts, so $T$ can also be learned by C-GNN. 
\begin{multline}
T(u,v)  := 
  \exists\,z_{1},z_{2}  (
    r_{1}(u,z_{1})  \wedge  r_{2}(z_{1},z_{2})  \wedge  r_{3}(z_{2},v)
  ) \\
   \wedge  
  \exists\,z_{1},z_{2} (
    r_{1}(u,z_{1})  \wedge  r_{4}(z_{1},z_{2})  \wedge  r_{5}(z_{2},v)).
\end{multline}

When the relations $r_{2}$ and $r_{4}$ are outgoing edges of the same node, we refer to this modified structure as \textit{U},
as shown in Figure \ref{fig:d} and the corresponding rule formula is given as follows:
\begin{multline}
U(u,v)  :=  \exists\,z_{1} r_{1}(u,z_{1})\wedge (\exists\,z_{2},z_{3}  (r_{2}(z_{1},z_{2}) \\ \wedge  r_{3}(z_{2},v)\wedge r_{4}(z_{1},z_{3}) \wedge  r_{5}(z_{3},v))
\end{multline}
In this case, if we do not add an additional variable to indicate that the edges $r_{2}$ and $r_{4}$ come from the same node, the logical rule structures $T$ and $U$ have the same CML expression. That is, C-GNN cannot distinguish between $U$ and $T$ under CML, so we have the following corollary:
\begin{corollary}
	\label{corollary:C-GNN}
	C-GNN cannot learn the logical rule structure $U$.
\end{corollary}

This limitation highlights the need for additional mechanisms to enhance the model's expressive power.
To address this issue, the labeling trick is introduced, a technique that augments the input representation to differentiate previously indistinguishable structures. In the following, we formally define the labeling trick and provide an example demonstrating how it enables C-GNN to overcome the limitation identified in Corollary \ref{corollary:C-GNN}.
\begin{definition}
  (\textbf{Labeling Trick}). 
Given a rule structure R(x,y) and an adjacency matrix $\textbf{A} \in \mathbb{R}^{ n \times n \times k} $ for the graph $\mathcal{G}$, we assign a constant to each node whose out-degree is greater than one. Construct a labeled tensor $L^{(S)} \in \mathbb{R}^{ n \times n \times d}$ to distinguish target nodes from non-target nodes. Concatenate $L^{(S)}$ and $\textbf{A}$ to get a new graph representation $A^{(S)} = [\textbf{A} \; L^{(S)}] \in \mathbb{R}^{ n \times n \times (k+d)}$, which allows C-GNN to learn the rules to distinguish them.
    
\end{definition}
For example, in Figure \ref{fig:all}, \textit{U} and \textit{T} cannot be effectively distinguished. When a constant $c$ is added to $z_1$ and a specific initial representation is given, the C-GNN can learn the structure of the rules $ U'(u,v):= r_{1}(u,c)\wedge (\exists\,z_{2},z_{3}  (r_{2}(c,z_{2})  \wedge  r_{3}(z_{2},v)\wedge r_{4}(c,z_{3})\wedge  r_{5}(z_{3},v))$. EL-GNN \cite{understanding} improves the rule learning ability by using the labeling trick based on C-GNN. However, when nodes are given unique labels, the model may over-rely on this ``artificial'' information during training, thereby ignoring other potential topological features or relationship clues, resulting in a decrease in the model's generalization ability.

Next, we introduce the concept of ``coverage'' \cite{Quinlan1990, MUGGLETON1994629} to assess the logical expressive power of these rules and explain why assigning constants to variables reduces coverage.

\begin{definition}
	\label{definition:cov}
   (\textbf{Coverage}). 
Given a knowledge graph $\mathcal{G}$, for the rule
$\varphi(x,y) := \exists z_{1},z_{2},\cdots (
  r_{1}(x,z_{1}) \wedge r_{2}(z_{1},z_{2}) \wedge \cdots \wedge r_{m}(z_{k},y))$,  $\{z_{1}, z_{2}, \cdots, z_{k}\} \in \mathcal{E}$,
we define coverage $Cov(\varphi)$ as the set of all pairs of entities $(x,y)$ that can satisfy the rule $\varphi(x,y)$ in $\mathcal{G}$. Formally,
\begin{equation}
\label{equation:cov}
Cov(\varphi) = \{(x,y) \in \mathcal{E} \times \mathcal{E} \mid \mathcal{G} \models \varphi(x,y)\}.
\end{equation}
Where $\mathcal{G} \models \varphi(x,y)$ means that there exist entities $\{z_{1},z_{2},\cdots \}$ in $\mathcal{G}$ that satisfy all the relational conditions in $\varphi(x,y)$. In other words, $Cov(\varphi)$ measures how many valid pairwise entities $(x,y)$ in $\mathcal{G}$ are ``covered'' by the rule $\varphi$. The greater the coverage, the stronger the generalizability of the rule.
\end{definition}

Regardless of the structure of the rule, as long as a variable is replaced by a constant in the rule, there will be a coverage reduction effect. That is to say, for any structure, if the original rule $\varphi(x,y)$ contains one or more variables, these variables determine which elements the rule can adapt to (larger coverage). After replacing the variable with a constant to obtain a new rule $\varphi{'}(x,y)$, it can only match the specific elements corresponding to the constant (smaller coverage). Therefore, it can be summarized as follows:
\begin{corollary}  
\label{corollary:coverage}
(\textbf{Adding Constants Restricts Coverage})
If we start from the original rules $\varphi(x,y, \cdots)$ and replace at least one of them with a constant c to obtain new rules $\varphi{'}(x,c, \cdots)$, the coverage of the new rules is less than that of the original rules. Formally expressed as: 
\begin{equation}
\label{equation:cov2}
Cov\Bigr(\varphi{'}(x,c, \cdots)\Bigr) \subseteq Cov\Bigr(\varphi(x,y, \cdots)\Bigr).
\end{equation}
Rules with constants (labels) become more specific, resulting in a smaller coverage and reduced generalization ability. 
\end{corollary}
According to corollary \ref{corollary:coverage}, we can conclude that
$Cov (EL\text{-}GNN)  \subseteq Cov(C\text{-}GNN)$

\section{Path neighbors enhanced GNN}

\subsection{Method}

To improve the logical expressive power of C-GNN without weakening its generalization ability, we propose PN-GNN, which uses a flexible plug-and-play approach without adding constant labels.
PN-GNN iteratively updates $L$ steps according to formula (\ref{nbfnet}) to obtain the pairwise entity representation ${\textbf{h}}_{v|u,q}^{(L)}$, and aggregates the representations of neighbor nodes along the path passing through the head and tail entities in the KG, leveraging these neighbors to enhance the structural expressive power of the GNN:
\begin{equation}
	\mathbf{h}_{ij} = \text{POOL}(\{\mathbf{h}_{w|u,q}^{(L)}|w \in P_{uv},d(u,w)=i,d(w,v)=j\})
\label{pool}
\end{equation}

The aggregated nodes are represented by $\textbf{h}_{ij}$. Among them, $i$ represents the distance from the head entity to the node $w$ to be aggregated, and $j$ represents the distance from the node $w$ to the tail entity $v$. $P_{uv}$ represents all paths with $u$ as the head source node and $v$ as the target node. $d(\cdot,\cdot)$ represents the distance between nodes. POOL is a pooling function that aggregates the representations of neighborhoods along the path with the same distance. It can be set to any commonly used function in GNN, such as \textit{max}, \textit{min}, \textit{mean}.

Using the aggregated representations $\textbf{h}_{ij}$ of the neighbors of nodes on the path, PN-GNN only needs to fuse the representation $\textbf{h}_{v|u,q}^{(L)}$ learned by C-GNN:
\begin{equation}
\begin{aligned}
&\textbf{h}_{d} = \bigoplus_{{i+j} \leqslant d}\text{MLP}(\textbf{h}_{ij})\\
&\textbf{h}_{v|u,q} = \textbf{h}_{d} \odot \textbf{h}_{v|u,q}^{(L)}
\end{aligned}
\label{hq}
\end{equation}

Formula (\ref{hq}) computes the node representation $\textbf{h}_{v|u,q}$ of each target tail entity $v$.
$\text{MLP}(\cdot)$ is a multi-layer perceptron. The operator $\bigoplus$ aggregates all neighbors in Equation (\ref{pool}), using operations like addition or concatenation. Similarly, $\odot$ fuses the C-GNN pairwise representation with the neighbor vector, also via addition or concatenation. To reduce computational overhead while ensuring the expressive power of PN-GNN, we select 1-hop and 2-hop neighbors along the path. Here, $\textbf{h}_{11}$ denotes the aggregation of neighbors with 1-hop, which means that neighbor node $w$ is 1-hop away from both the source node $u$ and the target node $v$. The 2-hop neighborhoods include $\textbf{h}_{12}$ and $\textbf{h}_{21}$, which indicate that $w$ is 1-hop from $u$ and 2-hop from $v$, or 2-hop from $u$ and 1-hop from $v$, respectively. 
\begin{figure}[t]
    \centering

    \subfigure[$T$]{
        \includegraphics[width=0.19\textwidth]{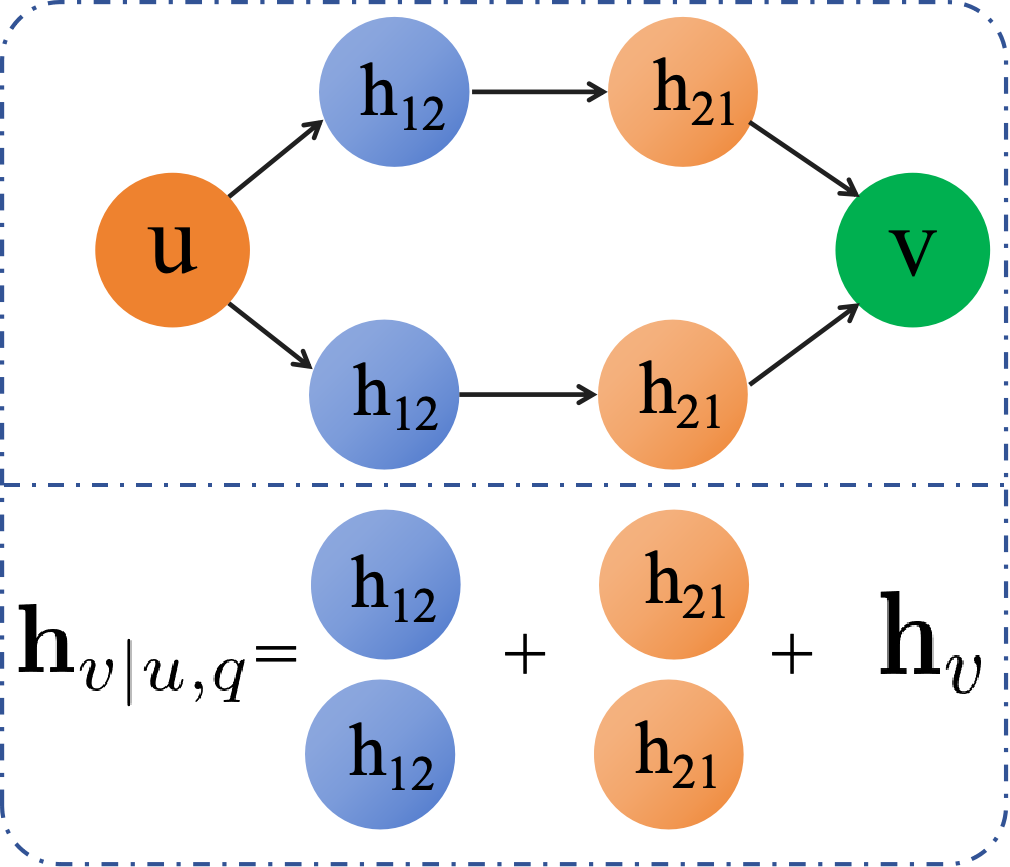}
        \label{fig:m2}
    }
    \subfigure[$U$]{
        \includegraphics[width=0.19\textwidth]{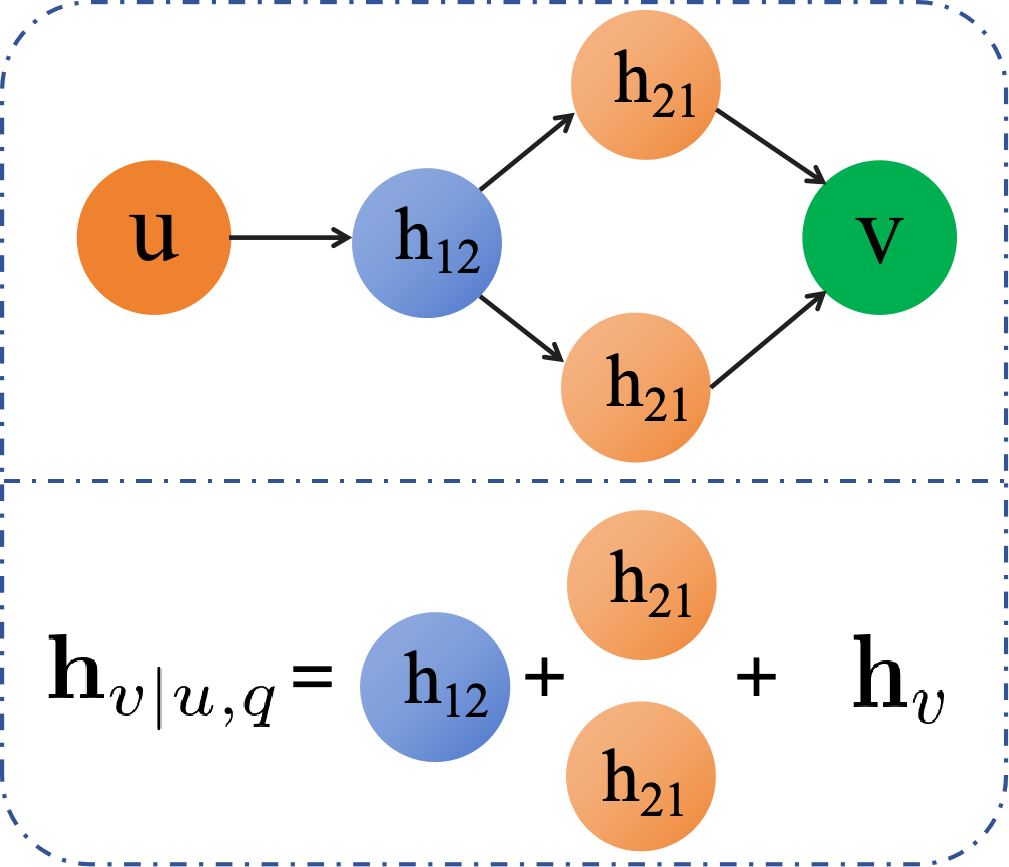}
        \label{fig:m2}
    }
    \caption{Example of the aggregation process}
    \label{fig:method}
\end{figure}
As shown in Figure \ref{fig:method},
Equation (\ref{hq}) now only involves path neighborhoods within two hops.

\begin{equation}
\begin{aligned}
&\textbf{h}_{d} = \text{MLP}_{11}(\textbf{h}_{11}) \odot \text{MLP}_{12}(\textbf{h}_{12}) \odot \text{MLP}_{21}(\textbf{h}_{21})\\
& \textbf{h}_{v|u,q} = \textbf{h}_{d} \odot \textbf{h}_{v|u,q}^{(L)}
\end{aligned}
\label{hq_2hop}
\end{equation}

The iterative update process of the pairwise-entity representation $\textbf{h}_{v|u,q}$ can be expressed by algorithm~\ref{alg_sec3}.
\begin{algorithm}[htbp]
	\caption{PN-GNN}
	\begin{algorithmic}[1]
		\label{alg_sec3}
		\REQUIRE Source node $u$, query relation $q$,  KG $\mathbb{G}$
		\ENSURE Node pair representation $\textbf{h}_{v|u,q}$, $v \in V$
		\FOR{$v \in V$}
		\STATE $\textbf{h}_{v|u,q}^{(0)} \leftarrow \text{INIT}(u,v,q)$
		\ENDFOR
		\FOR{$t \in [1,T]$}
		\FOR{$v \in V$}
		\STATE $\textbf{h}_{v|u,q}^{(t)} \leftarrow \text{C-GNN}(u,q,KG)$
		\ENDFOR
		\ENDFOR
		\STATE $\textbf{h}_{ij} = \text{POOL}(\{\textbf{h}_{w|u,q}^{(L)}|w \in P_{uv},d(u,w)=i,d(w,v)=j\})$
		\STATE $\textbf{h}_{d} = \bigoplus_{i,j \leqslant d}\text{MLP}(\textbf{h}_{ij})$
		\STATE $\textbf{h}_{v|u,q} = \textbf{h}_{d} \odot \textbf{h}_{v|u,q}^{(L)}$
		\RETURN $\textbf{h}_{v|u,q}$
	\end{algorithmic}
\end{algorithm}

After $L$ iterations, the model obtains the pairwise-entity representation $\textbf{h}_{v|u,q}$ for every entity $v$ in the KG. Similar to the C-GNN, PN-GNN directly uses this representation to compute the conditional probability of the tail entity $v$.
\begin{equation}
	p(v|u,q) = \sigma(\text{MLP}(\textbf{h}_{v|u,q}))
\end{equation}
Here, $\sigma(\cdot)$ is the sigmoid function, and $\text{MLP}(\cdot)$ denotes a multilayer perceptron. PN-GNN is trained by minimizing the negative log-likelihood over positive and negative triples:
\begin{equation}
L_{KG} = -logp(u,q,v) - \sum_{i=1}^{n}\frac{1}{n}log(1-p(u_{i}^{'},q,v_{i}^{'}))
\end{equation}
Here, $(u_{i}^{'}, q, v_{i}^{'})$ denotes a negative sample. Negative samples are constructed under the Partial Completeness Assumption (PCA) \cite{bordes2013translating,sun2019rotate} by randomly 
replacing the head (or tail) entity with an incorrect one from the knowledge graph, and $n$ denotes the number of negative samples. In order to balance logical expressive power and efficiency, we use a 2-hop PN-GNN.

\subsection{Logical Expressive Power of PN-GNN}

The matching of logical rule structures in KG is regarded as a binary logical classification task. Like C-GNN learns the pairs of nodes $(u,v)$ in KG, there is a FOL expression $\varphi(x,y)$, which contains two variables $x$ and $y$. If there is a pair of entities $(u,v)$ in the KG that satisfies $G \models \varphi(u,v)$, the tail entity is assigned to ``true''. In order to prove that the logical expressive of PN-GNN is more powerful than that of C-GNN, it is first proved that PN-GNN can learn all logical rule structures that C-GNN can learn. According to corollary \ref{corollary:C-GNN}, it is necessary to first prove that PN-GNN can learn the logical rule structure in CML in KG.

Let $\varphi(x)$ be a FOL expression in the CML of the KG. $\varphi$ can be decomposed into a series of sub-expressions, namely $sub(\varphi)=(\varphi_{1},\varphi_{2},...,\varphi_{L})$. If $\varphi_{k}$ is a sub-expression of $\varphi_{l}$, then $k \leqslant l$, and finally $\varphi = \varphi_{L}$. $\mathbf{h}_{v}^{(t)} \in \mathbb{R}^{L}$ is the representation learned by GNN through message passing, and $L$ is the total number of iterations. The iterative calculation of PN-GNN can be simplified to the following formula:
\begin{equation}
\mathbf{h}_{v}^{(t)} = \sigma(\mathbf{h}_{v}^{(t-1)}\mathbf{C} + \sum_{j=1}^{r}\sum_{u \in v}\mathbf{h}_{u}^{(t-1)}\mathbf{A}_{r} + \mathbf{b})
\label{eq:c-gnn2}
\end{equation}
\begin{equation}
\label{eq:pn-gnn2}
\mathbf{h}_{v} = \sigma(\mathbf{h}_{v}^{(L)} + \sum_{i+j = |d|}(\mathbf{h}_{ij}^{(L)}\mathbf{A}_{pn} + \mathbf{b}_{pn}))
\end{equation}

The formula (\ref{eq:c-gnn2}) represents the process of C-GNN iteratively calculating entity representation. Formula (\ref{eq:pn-gnn2}) represents the process of PN-GNN aggregating neighborhoods representations in the path, where $|d|$ represents the maximum length of the allowed path. For example, when $|d|=2$, the neighborhoods with a distance of 1-hop from the head and tail entities are aggregated. When $|d|=3$, the neighborhoods with a distance of 1-hop from the head entity and 2-hop from the tail entity and 2-hop from the head entity and 1-hop from the tail entity are aggregated. $\sigma = min(max(0,x),1)$ is the ReLU activation function.

Based on formula (\ref{eq:c-gnn2}) and (\ref{eq:pn-gnn2}), we can derive the following two lemmas:

\begin{lemma}
\label{lemma:PN-GNN-1}
If $\varphi(x)$ is a CML formula that can be learned by C-GNN, then PN-GNN can also learn $\varphi(x)$.
\end{lemma}

\begin{lemma}
 PN-GNN can learn the logical rule structure $U$.
\label{lemma:pn-gnn2}
\end{lemma}
Based on these lemmas, we show that PN-GNN can learn the logical rule structure $U$,  while C-GNN cannot (see Corollary~\ref{corollary:C-GNN}). This leads to the conclusion that PN-GNN has stronger logical expressiveness than C-GNN. The theorem is stated below (see Appendix A.2 for complete proofs):
\begin{theorem}
\label{theorem:PN-GNN-2}
PN-GNN has strictly more powerful logical expressiveness than C-GNN.
\end{theorem}
The expressive power of PN-GNN depends on both its architecture and propagation range (number of hops). As the propagation range increases, its logical expressiveness improves. The following results formalize this relationship:

\begin{lemma}
(k+1)-hop can represent all logical expressions that k-hop can learn.
\label{lemma:kk1}
\end{lemma}
 
\begin{lemma}
    (k+1)-hop can learn logical rules that cannot be expressed by k-hop.
    \label{lemma:kk2}
\end{lemma}
Based on Lemmas \ref{lemma:kk1} and \ref{lemma:kk2}, we obtain the following theorem (see Appendix A.2 for the complete proof):
\begin{theorem}
\label{theorem:PN-GNN-3}
For PN-GNN, the logical expressive power of (k+1)-hop is strictly more powerful than that of k-hop.
\end{theorem}
Although increasing the number of hops enhances the model’s logical expressive capability, it also reduces computational efficiency. Therefore, to balance expressiveness and efficiency, we set k = 2.

\section{Experiments}
 
\subsection{Experiments Setting}
\paragraph{Datasets} To evaluate the logical expressive power of PN-GNN, we conducted experiments on two real-world datasets (WN18RR and FB15K237) in the transductive setting and their inductive variants (v1-v4), as well as six synthetic datasets provided by Qiu et al. (2024). We also constructed two additional synthetic datasets, $T_{\text{label}}$ and $U_{\text{label}}$, to examine how labeling strategies affect generalization and ensure predictions rely on logical structure rather than specific labels. Details on dataset construction, statistics, hyperparameters, and evaluation metrics are provided in Appendix B.1.

\paragraph{Baselines} We compare PN-GNN with several baselines, including R-GNN-based methods, C-GNN-based methods, and labeling strategy-based EL-GNN. In addition, we use representation learning, path-based, and GNN-based methods as baselines in both inference and inductive settings. See Appendix B.4 for more details.

\subsection{Experiments On Synthetic Datasets}
Results on synthetic datasets are shown in Table ~\ref{sec3:construct_results}. The results based on the R-GNN method come from Qiu et al. (2024). The results of the EL-GNN method are obtained by using NBFNet as the backbone and carefully searching for parameters on the validation set using the Ray \cite{liaw2018tune}. Through the analysis of the results, we can observe that:

(1)Theoretical analysis indicates that the logical expressive of C-GNN is strictly more powerful than that of R-GNN. Experimental results validate this conclusion: methods based on C-GNN, such as NBFNet, RED-GNN, EL-GNN, and the proposed PN-GNN, achieved 100\% accuracy on the $C_{3}$, $C_{4}$, $I_{1}$, $I_{2}$, and $T$. In contrast, the performance of R-GNN confirms that it is nearly incapable of learning these logical rules. Notably, PN-GNN also achieved 100\% accuracy on all datasets except $U$ and the two datasets ($T_{label}$ and $U_{label}$) that we constructed, demonstrating that its logical expressive power is comparable to that of C-GNN and that it can learn the logical rules in CML.

(2) EL-GNN and PN-GNN exhibit strictly more powerful logical expressive compared to C-GNN. Analysis indicates that C-GNN does not distinguish between $U$ and $T$, making it impossible to learn the logical rule structures in $U$. In contrast, EL-GNN, leveraging a labeling strategy, achieved the best performance on the $U$ dataset, with a 21.6\% improvement over C-GNN. Similarly, PN-GNN also outperformed C-GNN, achieving a 15.8\% improvement. These results demonstrate that PN-GNN enhances the logical expressive power of C-GNN.

(3) We demonstrate the impact of constant labels on EL-GNN using the $T_{label}$ and $U_{label}$ datasets. The performance of EL-GNN significantly degrades on both datasets, primarily because the model attaches fixed labels to nodes during training and learns node representations based on these constant labels. However, during testing, due to encountering different constant labels, the learned rules fail to fully cover all cases, leading to a decline in performance. This indicates that the labeling trick undermines the model’s generalization ability. In particular, for the $T_{label}$ dataset, where labels are not required, it is affected by the label vector during training, resulting in reliance on labels for prediction during testing, which leads to a significant drop in performance. In contrast, PN-GNN exhibits a much stronger performance: on the $T_{label}$ dataset, it performs on par with NBFNet and far outperforms EL-GNN, and on the $U_{label}$ dataset, PN-GNN not only surpasses EL-GNN but also outperforms NBFNet. This demonstrates that our approach can effectively alleviate the negative impact of constant labels on model generalization.

\begin{table}[ht]
	\centering
		\begin{tabularx} {\linewidth}{cYYYYYYcc}
			\toprule 
            	Method & $C_{3}$ & $C_{4}$ & $I_{1}$ &  $I_{2}$ & $T$ &  $U$ &$T_{label}$&$U_{label}$\\
			\midrule 
		R-GCN & 1.6 & 3.1& 4.4& 2.4&6.7&1.4 &-&-\\
			CompGCN&1.6&2.1&5.3&3.9&6.7&2.7 &-&-\\
            \midrule
			NBFNet&100&100&100&100&100&54.1&60.0&56.8 \\
			EL-GNN&100&100&100&100&100&75.7&22.0&59.5 \\
			PN-GNN&100&100&100&100&100&69.9&60.0&68.9 \\
			\bottomrule 
		\end{tabularx}
        \caption{Results on synthetic datasets (Hits@1). }
		\label{sec3:construct_results}
\end{table}

\begin{table*}[t]
	\centering
		
        	\begin{tabular}{ccccccccc}
			\toprule 
			\multirow{2}{*}{Method}
			&\multicolumn{4}{c}{\bf{FB15K237}} & \multicolumn{4}{c}{\bf{WN18RR}}\\
			&MRR&Hits@1&Hits@3&Hits@10&MRR&Hits@1&Hits@3&Hits@10\\
			\midrule 
			TransE&0.294&-&-&46.5&0.226&-&40.3&53.2\\
			RotatE&0.338&24.1&37.5&53.3&0.476&42.8&49.2&57.1\\
			HAKE&0.341&24.3&37.5&53.5&0.496&45.1&51.3&58.2\\
			RotH&0.344&24.6&38.0&53.5&0.495&44.9&51.4&58.6\\
			ComplE+RP&0.388&29.8&42.5&56.8&0.488&44.3&50.5&57.8\\
			ConE&0.345&24.7&38.1&54.0&0.496&45.3&51.5&57.9\\
			NeuralLP&0.240&-&-&36.2&0.435&37.1&43.4&56.6\\
			DRUM&0.343&25.5&37.8&51.6&0.486&42.5&51.3&58.6\\
            \midrule 
			R-GCN&0.273&18.2&30.3&45.6&0.402&34.5&43.7&49.4\\
			CompGCN&0.355&26.4&39.0&53.5&0.479&44.3&49.4&54.6\\
			NBFNet&0.415&32.1&45.4&59.9&0.551&49.7&57.3&66.6\\
			RED-GNN&0.374&28.3&-&55.8&0.533&48.5&-&62.4\\
			A*Net&0.411&32.1&45.3&58.6&0.549&49.5&57.3&65.9\\
			EL-GNN&\bf{0.421}&\bf{33.2}&45.9&59.8&\bf{0.555}&\bf{49.9}&\bf{57.8}&66.4\\
			PN-GNN&\bf{0.423}&\bf{33.1}&\bf{46.5}&\bf{60.2}&\bf{0.555}&\bf{49.9}&\bf{57.9}&\bf{66.9}\\
			\bottomrule
		\end{tabular}
        \caption{Results for transductive setting on real-world datasets. The best results are shown in bold.}
		\label{sec3:transductive_results}
\end{table*}

\subsection{Transductive results}

The results of the transductive setting are shown in Table ~\ref{sec3:transductive_results}. 
Across all evaluation metrics, C-GNN based methods surpass those based on representation learning, paths, and R-GNNs. This outcome aligns with C-GNN theory, showing that C-GNN can learn and leverage logical rule structures. They also indicate that, although GNNs capture more graph structures than path-based methods, the ability to learn logical rule structures is crucial for KG reasoning. PN-GNN and EL-GNN further improve upon C-GNN, demonstrating that enhancing the representation of logical rule structures leads to better inference performance. On the real-world dataset FB15K237, which contains $U$ and $T$ structures, EL-GNN and PN-GNN outperform C-GNN. Notably, the node out-degree $d$ is a sensitive hyperparameter for EL-GNN and has a great impact on performance \cite{understanding}, whereas PN-GNN does not require this parameter.

\begin{table}[t]
	\centering
\begin{tabularx} {\linewidth}{c|YYYYYYYY}
			\toprule 
			\multirow{2}{*}{Method}
			&\multicolumn{4}{c}{\bf{FB15K237}} & \multicolumn{4}{c}{\bf{WN18RR}}\\
			&v1&v2&v3&v4&v1&v2&v3&v4\\
			\midrule 
			NeuralLP & 52.9 & 58.9 & 52.9 & 55.9 & 74.4 & 68.9 & 46.3 & 67.1 \\
			DRUM & 52.9 & 58.7 & 52.9 & 55.9 & 74.4 & 68.9 & 46.2 & 67.1 \\
			RuleN & 49.8 & 77.8 & 87.7 & 85.6 & 80.9 & 78.2 & 53.4 & 71.6\\
			GraIL & 64.3 & 81.8 & 82.8 & 89.3 & 82.4 & 78.7 & 58.4 & 73.4 \\
			CoMPILE & 67.6 & 82.9& 84.7& 87.4& 83.6& 79.8& 60.7& 75.5\\
			RED-GNN & 65.7& 82.9 & 81.1 & 89.9 & 86.7 & 83.1 &70.8 & 78.6\\
			NBFNet & 83.5 & 94.9 & 95.2 & 96.1 & 94.8 & 90.1 & 89.3 & 89.1\\
			PN-GNN &\bf{85.1}&\bf{95.8}&\bf{96.2}&\bf{96.5}&\bf{95.2}&\bf{90.8} & \bf{89.8}&\bf{89.3}\\
			\bottomrule
		\end{tabularx}
		\caption{Results for inductive setting (evaluated with Hits@10). The best results are shown in bold.}
		\label{sec3:inductive_results}
\end{table}

\begin{table}[t]
	\centering
\begin{tabularx} {\linewidth}{c|ccYY}
			\toprule 
			{Method} & { $T$} & {  $U$} & {FB15K237v1} & {WN18RRv1}\\
			\midrule
			NBFNet&100&54.1&17.1&60.1\\
			PN-GNN&100&69.9&22.1&62.5\\
			$\text{PN-GNN}_{11}$&100&48.7&20.2&62.8\\
			$\text{PN-GNN}_{12-21}$&100&69.9&18.5&62.5\\
			\bottomrule 
		\end{tabularx}
        \caption{Ablation study results on dataset using the Hits@1.}
		\label{sec3:ablation}
\end{table}
\subsection{Inductive results}

Table~\ref{sec3:inductive_results} presents the experimental results under the inductive setting. 
EL-GNN relies on entity representations to annotate entities, rendering it incapable of handling unseen entities in the inductive setting. 
In the inductive setting, models must infer unseen entities from existing relationships, directly reflecting their logical expressiveness. 
Experimental results show that C-GNN-based methods, such as NBFNet and RED-GNN, outperform path-based approaches. 
Building on NBFNet, PN-GNN achieves further improvements across datasets, verifying the effectiveness of its enhanced logical expressiveness in inductive scenarios. 
We also conduct experiments to test the effectiveness of PN-GNN and provide detailed comparisons in Appendix B.5.

\subsection{Ablation Study}

To examine the effect of neighborhood hops on PN-GNN's performance, we conducted ablation studies on $T$, $U$, FB15K237 v1, and WN18RR v1, comparing $\text{PN-GNN}_{11}$ (using only 1-hop) and $\text{PN-GNN}_{12-21}$ (using only 2-hop). Results are shown in Table~\ref{sec3:ablation}.
C-GNN is sufficient for the $T$ dataset, with all models achieving 100\% accuracy. 
For $U$, $\text{PN-GNN}_{12-21}$ outperforms NBFNet by 15.8\%, consistent with Lemma~\ref{lemma:pn-gnn2}, as distinguishing the logical rules in $U$ mainly relies on 3-hop neighborhoods.
On FB15K237 v1, PN-GNN and its variants all outperform NBFNet, with the best improvement reaching 5\%, demonstrating the benefits of improved logical expressiveness in complex real-world datasets. 
On WN18RR v1, $\text{PN-GNN}_{11}$ achieves the highest performance, outperforming NBFNet by 2.7\%. This suggests that for datasets with fewer relations and predominantly short-path logical structures, leveraging only 1-hop neighborhoods provides more relevant information than considering longer paths, which may introduce noise.

\section{Conclusion }
In this work, we analyze the logical expressiveness of GNNs in KG reasoning. We highlight the limitations of the labeling trick—commonly used to boost link prediction—which can hinder model generalization. To address this, we propose Path-Neighbor enhanced GNN (PN-GNN), which improves the logical expressive power of GNN by aggregating node-neighbor embeddings on the reasoning paths and avoids the generalization issues of the labeling trick, while still supporting inductive settings. 
PN-GNN faces increased computational costs in multi-hop scenarios as resource demands grow with hops. Future work will aim to enhance both efficiency and expressiveness for larger, more complex tasks.

%
\section*{Acknowledgements}
This work is particularly supported by the National Key Research and Development Program of China (No. 2022YFB3104103), the National Natural Science Foundation of China (No. 62302507), the Provincial Natural Science Foundation of Hunan (No. 2024JJ7249, 2025JJ70302), and the National First-Class Undergraduate Major (Network Engineering) Fund.

\bigskip

\bibliography{aaai2026}

@article{Quinlan1990,
  title={Learning logical definitions from relations},
  author={J. Ross Quinlan},
  journal={Machine Learning},
  year={1990},
  volume={5},
  pages={239-266},
  url={https://api.semanticscholar.org/CorpusID:6746439}
}

@article{MUGGLETON1994629,
title = {Inductive Logic Programming: Theory and methods},
journal = {The Journal of Logic Programming},
volume = {19-20},
pages = {629-679},
year = {1994},
note = {Special Issue: Ten Years of Logic Programming},
issn = {0743-1066},
doi = {https://doi.org/10.1016/0743-1066(94)90035-3},
url = {https://www.sciencedirect.com/science/article/pii/0743106694900353},
author = {Stephen Muggleton and Luc {de Raedt}},
}

@inproceedings{LabelingTrick,
  author       = {Muhan Zhang and
                  Pan Li and
                  Yinglong Xia and
                  Kai Wang and
                  Long Jin},

  title        = {Labeling Trick: {A} Theory of Using Graph Neural Networks for Multi-Node
                  Representation Learning},
  booktitle    = {NeurIPS},
  pages        = {9061--9073},
  year         = {2021},

}

@inproceedings{Srinivasan020,
  author       = {Balasubramaniam Srinivasan and
                  Bruno Ribeiro},
  title        = {On the Equivalence between Positional Node Embeddings and Structural
                  Graph Representations},
  booktitle    = {ICLR 2020},
 
  year         = {2020},

}

@article{wu2020comprehensive,
	title={A comprehensive survey on graph neural networks},
	author={Wu, Zonghan and Pan, Shirui and Chen, Fengwen and Long, Guodong and Zhang, Chengqi and Philip, S Yu},
	journal={IEEE transactions on neural networks and learning systems},
	volume={32},
	number={1},
	pages={4--24},
	year={2020},
	publisher={IEEE}
}

@inproceedings{velivckovic2017graph,
	author       = {Petar Velickovic and
	Guillem Cucurull and
	Arantxa Casanova and
	Adriana Romero and
	Pietro Li{\`{o}} and
	Yoshua Bengio},
	title        = {Graph Attention Networks},
	booktitle    = {6th International Conference on Learning Representations, {ICLR} 2018,
	Vancouver, BC, Canada, April 30 - May 3, 2018, Conference Track Proceedings},
	year         = {2018},
}

@inproceedings{hamilton2017inductive,
	author       = {William L. Hamilton and
	Zhitao Ying and
	Jure Leskovec},
	title        = {Inductive Representation Learning on Large Graphs},
	booktitle    = {NeurIPS 2017},
	pages        = {1024--1034},
	year         = {2017},
}

@inproceedings{Barcelo00O22,
  author       = {Pablo Barcel{\'{o}} and
                  Mikhail Galkin and
                  Christopher Morris and
                  Miguel A. Romero Orth},
  title        = {Weisfeiler and Leman Go Relational},
  booktitle    = {Learning on Graphs Conference, LoG 2022, 9-12 December 2022, Virtual
                  Event},
  series       = {Proceedings of Machine Learning Research},
  volume       = {198},
  pages        = {46},
  publisher    = {{PMLR}},
  year         = {2022},
}

@inproceedings{xu2018powerful,
	author       = {Keyulu Xu and
	Weihua Hu and
	Jure Leskovec and
	Stefanie Jegelka},
	title        = {How Powerful are Graph Neural Networks?},
	booktitle    = {7th International Conference on Learning Representations, {ICLR} 2019,
	New Orleans, LA, USA, May 6-9, 2019},
	year         = {2019},
}

@inproceedings{schlichtkrull2018modeling,
	title={Modeling relational data with graph convolutional networks},
	author={Schlichtkrull, Michael and Kipf, Thomas N and Bloem, Peter and Van Den Berg, Rianne and Titov, Ivan and Welling, Max},
	booktitle={ESWC 2018},
	pages={593--607},
	year={2018},
	organization={Springer}
}

@inproceedings{morris2019weisfeiler,
  author       = {Christopher Morris and
                  Martin Ritzert and
                  Matthias Fey and
                  William L. Hamilton and
                  Jan Eric Lenssen and
                  Gaurav Rattan and
                  Martin Grohe},
  title        = {Weisfeiler and Leman Go Neural: Higher-Order Graph Neural Networks},
  booktitle    = {AAAI 2019},
  pages        = {4602--4609},
  publisher    = {{AAAI} Press},
  year         = {2019},
  url          = {https://doi.org/10.1609/aaai.v33i01.33014602},
  doi          = {10.1609/AAAI.V33I01.33014602},
  bibsource    = {dblp computer science bibliography, https://dblp.org}
}

@inproceedings{SatoYK21,
  author       = {Ryoma Sato and
                  Makoto Yamada and
                  Hisashi Kashima},

  title        = {Random Features Strengthen Graph Neural Networks},
  booktitle    = {Proceedings of the 2021 {SIAM} International Conference on Data Mining,
                  {SDM} 2021, Virtual Event, April 29 - May 1, 2021},
  pages        = {333--341},
  publisher    = {{SIAM}},
  year         = {2021},

}

@inproceedings{you2021identity,
	title={Identity-aware graph neural networks},
	author={You, Jiaxuan and Gomes-Selman, Jonathan M and Ying, Rex and Leskovec, Jure},
	booktitle={AAAI 2021},
	volume={35},
	number={12},
	pages={10737--10745},
	year={2021}
}

@inproceedings{vashishth2019composition,
	author       = {Shikhar Vashishth and
	Soumya Sanyal and
	Vikram Nitin and
	Partha P. Talukdar},
	title        = {Composition-based Multi-Relational Graph Convolutional Networks},
	booktitle    = {ICLR 2020},
	year         = {2020},
}

@article{zhu2021neural,
	title={Neural bellman-ford networks: A general graph neural network framework for link prediction},
	author={Zhu, Zhaocheng and Zhang, Zuobai and Xhonneux, Louis-Pascal and Tang, Jian},
	journal={NeurIPS 2021},
	volume={34},
	pages={29476--29490},
	year={2021}
}

@inproceedings{zhu2024net,
	author       = {Zhaocheng Zhu and
	Xinyu Yuan and
	Michael Galkin and
	Louis{-}Pascal A. C. Xhonneux and
	Ming Zhang and
	Maxime Gazeau and
	Jian Tang},
	title        = {A*Net: {A} Scalable Path-based Reasoning Approach for Knowledge Graphs},
	booktitle    = {NeurIPS},
	year         = {2023},
}

@inproceedings{zhang2022knowledge,
	title={Knowledge graph reasoning with relational digraph},
	author={Zhang, Yongqi and Yao, Quanming},
	booktitle={Proceedings of the ACM web conference 2022},
	pages={912--924},
	year={2022}
}

@article{bordes2013translating,
	title={Translating embeddings for modeling multi-relational data},
	author={Bordes, Antoine and Usunier, Nicolas and Garcia-Duran, Alberto and Weston, Jason and Yakhnenko, Oksana},
	journal={NeurIPS},
	volume={26},
	year={2013}
}

@inproceedings{bai2021modeling,
 author       = {Yushi Bai and
                  Zhitao Ying and
                  Hongyu Ren and
                  Jure Leskovec},
  title        = {Modeling Heterogeneous Hierarchies with Relation-specific Hyperbolic
                  Cones},
  booktitle    = {Advances in Neural Information Processing Systems 34: Annual Conference
                  on Neural Information Processing Systems 2021, NeurIPS 2021, December
                  6-14, 2021, virtual},
  pages        = {12316--12327},
  year         = {2021},
  url          = {https://proceedings.neurips.cc/paper/2021/hash/662a2e96162905620397b19c9d249781-Abstract.html},
  bibsource    = {dblp computer science bibliography, https://dblp.org}
}

@misc{liaw2018tune,
      title={Tune: A Research Platform for Distributed Model Selection and Training}, 
      author={Richard Liaw and Eric Liang and Robert Nishihara and Philipp Moritz and Joseph E. Gonzalez and Ion Stoica},
      year={2018},
      eprint={1807.05118},
      archivePrefix={arXiv},
      primaryClass={cs.LG},
      url={https://arxiv.org/abs/1807.05118}, 
}

@article{corso2020principal,
	title={Principal neighbourhood aggregation for graph nets},
	author={Corso, Gabriele and Cavalleri, Luca and Beaini, Dominique and Li{\`o}, Pietro and Veli{\v{c}}kovi{\'c}, Petar},
	journal={Advances in Neural Information Processing Systems},
	volume={33},
	pages={13260--13271},
	year={2020}
}

@inproceedings{sun2019rotate,
	author       = {Zhiqing Sun and
	Zhi{-}Hong Deng and
	Jian{-}Yun Nie and
	Jian Tang},
	title        = {RotatE: Knowledge Graph Embedding by Relational Rotation in Complex
	Space},
	booktitle    = {ICLR 2019},
	year         = {2019},
}

@inproceedings{chami2020low,
	author       = {Ines Chami and
	Adva Wolf and
	Da{-}Cheng Juan and
	Frederic Sala and
	Sujith Ravi and
	Christopher R{\'{e}}},
	title        = {Low-Dimensional Hyperbolic Knowledge Graph Embeddings},
	booktitle    = {Proceedings of the 58th Annual Meeting of the Association for Computational
	Linguistics, {ACL} 2020, Online, July 5-10, 2020},
	pages        = {6901--6914},
	year         = {2020},
}

@inproceedings{zhang2020learning,
	title={Learning hierarchy-aware knowledge graph embeddings for link prediction},
	author={Zhang, Zhanqiu and Cai, Jianyu and Zhang, Yongdong and Wang, Jie},
	booktitle={Proceedings of the AAAI conference on artificial intelligence},
	volume={34},
	number={03},
	pages={3065--3072},
	year={2020}
}

@inproceedings{dettmers2018convolutional,
  author       = {Tim Dettmers and
                  Pasquale Minervini and
                  Pontus Stenetorp and
                  Sebastian Riedel},
  title        = {Convolutional 2D Knowledge Graph Embeddings},
  booktitle    = {AAAI },
  pages        = {1811--1818},
 
  year         = {2018},

}

@inproceedings{yang2014embedding,
	author       = {Bishan Yang and
	Wen{-}tau Yih and
	Xiaodong He and
	Jianfeng Gao and
	Li Deng},
	editor       = {Yoshua Bengio and
	Yann LeCun},
	title        = {Embedding Entities and Relations for Learning and Inference in Knowledge
	Bases},
	booktitle    = {3rd International Conference on Learning Representations, {ICLR} 2015,
	San Diego, CA, USA, May 7-9, 2015, Conference Track Proceedings},
	year         = {2015},
}

@inproceedings{trouillon2016complex,
	title={Complex embeddings for simple link prediction},
	author={Trouillon, Th{\'e}o and Welbl, Johannes and Riedel, Sebastian and Gaussier, {\'E}ric and Bouchard, Guillaume},
	booktitle={International conference on machine learning},
	pages={2071--2080},
	year={2016},
	organization={PMLR}
}

@inproceedings{meilicke2018fine,
	title={Fine-grained evaluation of rule-and embedding-based systems for knowledge graph completion},
	author={Meilicke, Christian and Fink, Manuel and Wang, Yanjie and Ruffinelli, Daniel and Gemulla, Rainer and Stuckenschmidt, Heiner},
	booktitle={The Semantic Web--ISWC 2018: 17th International Semantic Web Conference, Monterey, CA, USA, October 8--12, 2018, Proceedings, Part I 17},
	pages={3--20},
	year={2018},
	organization={Springer}
}

@inproceedings{yang2017differentiable,
	author       = {Fan Yang and
	Zhilin Yang and
	William W. Cohen},
	editor       = {Isabelle Guyon and
	Ulrike von Luxburg and
	Samy Bengio and
	Hanna M. Wallach and
	Rob Fergus and
	S. V. N. Vishwanathan and
	Roman Garnett},
	title        = {Differentiable Learning of Logical Rules for Knowledge Base Reasoning},
	booktitle    = {Advances in Neural Information Processing Systems 30: Annual Conference
	on Neural Information Processing Systems 2017, December 4-9, 2017,
	Long Beach, CA, {USA}},
	pages        = {2319--2328},
	year         = {2017},
}

@article{sadeghian2019drum,
	title={Drum: End-to-end differentiable rule mining on knowledge graphs},
	author={Sadeghian, Ali and Armandpour, Mohammadreza and Ding, Patrick and Wang, Daisy Zhe},
	journal={Advances in Neural Information Processing Systems},
	volume={32},
	year={2019}
}

@inproceedings{teru2020inductive,
	title={Inductive relation prediction by subgraph reasoning},
	author={Teru, Komal and Denis, Etienne and Hamilton, Will},
	booktitle={ICML 2020},
	pages={9448--9457},
	year={2020},
	organization={PMLR}
}

@inproceedings{toutanova2015representing,
	title={Representing text for joint embedding of text and knowledge bases},
	author={Toutanova, Kristina and Chen, Danqi and Pantel, Patrick and Poon, Hoifung and Choudhury, Pallavi and Gamon, Michael},
	booktitle={Proceedings of the 2015 conference on empirical methods in natural language processing},
	pages={1499--1509},
	year={2015}
}

@inproceedings{mai2021communicative,
	title={Communicative message passing for inductive relation reasoning},
	author={Mai, Sijie and Zheng, Shuangjia and Yang, Yuedong and Hu, Haifeng},
	booktitle={AAAI 2021},
	volume={35},
	number={5},
	pages={4294--4302},
	year={2021}
}

@article{goldberg1993heuristic,
title = {A heuristic improvement of the Bellman-Ford algorithm},
journal = {Applied Mathematics Letters},
volume = {6},
number = {3},
pages = {3-6},
year = {1993},
issn = {0893-9659},
author = {Andrew V. Goldberg and Tomasz Radzik},
}

@inproceedings{barcelo2020logical,
	title={The logical expressiveness of graph neural networks},
	author={Barcel{\'o}, Pablo and Kostylev, Egor V and Monet, Mikael and P{\'e}rez, Jorge and Reutter, Juan and Silva, Juan-Pablo},
	booktitle={ICLR 2020},
	year={2020}
}

@inproceedings{ATheory,
  author       = {Xingyue Huang and
                  Miguel Romero and
                  {\.I}smail {\.I}lkan Ceylan and
                  Pablo Barcel{\'{o}}},
  title        = {A Theory of Link Prediction via Relational Weisfeiler-Leman on Knowledge
                  Graphs},
  booktitle    = {NeurIPS},
  year         = {2023},
  
}

@article{cai1992optimal,
	title={An optimal lower bound on the number of variables for graph identification},
	author={Cai, Jin-Yi and F{\"u}rer, Martin and Immerman, Neil},
	journal={Combinatorica},
	volume={12},
	number={4},
	pages={389--410},
	year={1992},
	publisher={Springer}
}

@inproceedings{AbboudCGL21,
author       = {Ralph Abboud and
                  {\.I}smail {\.I}lkan Ceylan and
                  Martin Grohe and
                  Thomas Lukasiewicz},
  title        = {The Surprising Power of Graph Neural Networks with Random Node Initialization},
  booktitle    = {IJCAI 2021},
  pages        = {2112--2118},
  year         = {2021},
  url          = {https://doi.org/10.24963/ijcai.2021/291},
  doi          = {10.24963/IJCAI.2021/291},
  bibsource    = {dblp computer science bibliography, https://dblp.org}
}

@misc{otto2023graded,
      title={Graded modal logic and counting bisimulation}, 
      author={Martin Otto},
      year={2023},
      eprint={1910.00039},
      archivePrefix={arXiv},
      primaryClass={math.LO},
      url={https://arxiv.org/abs/1910.00039}, 
}

@inproceedings{kingma2014adam,
	author       = {Diederik P. Kingma and
	Jimmy Ba},
	editor       = {Yoshua Bengio and
	Yann LeCun},
	title        = {Adam: {A} Method for Stochastic Optimization},
	booktitle    = {3rd International Conference on Learning Representations, {ICLR} 2015,
	San Diego, CA, USA, May 7-9, 2015, Conference Track Proceedings},
	year         = {2015},
}

@misc{ba2016layer,
      title={Layer Normalization}, 
      author={Jimmy Lei Ba and Jamie Ryan Kiros and Geoffrey E. Hinton},
      year={2016},
      eprint={1607.06450},
      archivePrefix={arXiv},
      primaryClass={stat.ML},
      url={https://arxiv.org/abs/1607.06450}, 
}

@inproceedings{understanding,
  author       = {Haiquan Qiu and
                  Yongqi Zhang and
                  Yong Li and
                  Quanming Yao},
  title        = {Understanding Expressivity of {GNN} in Rule Learning},
  booktitle    = {ICLR 2024},
  year         = {2024},
  url          = {https://openreview.net/forum?id=43cYe4oogi},
  bibsource    = {dblp computer science bibliography, https://dblp.org}
}

\clearpage
\section*{Appendix}
\subsection*{A. Proof}
\label{appendix:proof}
\subsubsection{A.1 The logical Expressiveness of C-GNN}


As a supplement to the main text, this section first introduces the formal definition of CML in KG \cite{ATheory,understanding} to facilitate the proofs, and then presents the formal proofs of Theorems 1 and 2, which characterize the logical expressiveness of C-GNN.

\begin{manualtheorem}{Definition}{A.1}
Let $\mathcal{G},v \models P_{i}$ represent that entity $v$ in KG satisfies the unary logical predicate $P_{i}(x)$. CML in KG satisfies:
\begin{itemize}
\item If $\varphi(x) = \top$, then $\mathcal{G},v \models \varphi$ holds if $v$ is an entity in KG $\mathcal{G}$.
\item $\omega(x)=\varphi(x) \wedge \psi(x)$, $\mathcal{G},v \models \omega$ holds if and only if $\mathcal{G},v \models \varphi$, $\mathcal{G},v \models \psi$ both hold.
\item If $\varphi(x) = \neg\phi(x)$, $\mathcal{G},v \models \varphi$ holds if and only if $\mathcal{G},v \not\models \phi$ holds.
\item $\varphi(x)=\exists^{N \geq n}z(\phi(x,z) \wedge r(z,y))$, $\mathcal{G},v \models \varphi$ holds if and only if there are at least $n$ groups of entities satisfying $\{u|u \in N_{r}(v)\}$ and $\mathcal{G},u \models \phi$ holds.

\end{itemize}
\end{manualtheorem}


Theorems 1 and 2 are adapted from Theorem 3.2 in \cite{understanding} and Theorem 5.3 in \cite{ATheory}. For implementation-level details, we refer the reader to the original sources. We restate these theorems here for convenience:
\begin{manualtheorem}{Theorem}{1 (Restated)}
	\label{theorem:CGNN1}
C-GNN can learn every CML-based formula $\varphi(x)$ by leveraging its entity representations.
\end{manualtheorem}


\begin{proof}[proof of Theorem 1]
    \label{proof:cgnn1}
Let $\varphi(x)$ be a formula in CML. We decompose $\varphi$ into a series of sub-formulas 
$\text{sub}[\varphi] = (\varphi_1, \varphi_2, \dots, \varphi_L)$ where $\varphi_k$ is a sub-formula of $\varphi_\ell$ if $k \leq \ell$ and $\varphi = \varphi_L$. Assume $\mathbf{h}_v^{(i)} \in \mathbb{R}^L$ is the representation of C-GNN, $v \in \mathcal{V}, i = 1 \dots L$.
Then we analysis based on the following equation:

\[
\mathbf{h}_v^{(i)} = \sigma \left( \mathbf{h}_v^{(i-1)} \mathbf{C} + \sum_{j=1}^{r} \sum_{u \in \mathcal{N}_{r_j}(v)} \mathbf{h}_u^{(i-1)} \mathbf{A}_{r_j} + \mathbf{b} \right).
\]
To capture the connection between logical satisfaction and node representations in C-GNN, we introduce the following definition.
\begin{manualtheorem}{Definition}{A.2}
\label{definition:appendix}
    We set the initial representation  $\mathbf{h}_v^{(0)} = (t_1, t_2,
    \dots, t_n)$ has $t_{\ell} = 1$ if the sub-formula $\varphi_{\ell} = P_{\ell}(x)$ is satisfied at $v$, and $t_{\ell} = 0$ otherwise.
And for every $\ell \leq i \leq L$,  $
(h_v^{(i)})_{\ell} = 1 \quad \text{if } G, v \models \varphi_{\ell}.$
If $\varphi_{\ell}(x) = P_{\ell}(x)$ where $P_{\ell}$ is a unary predicate, $\mathbf{C}_{\ell\ell} = 1$ and all other values set to 0. We have $(h_v^{(0)})_{\ell} = 1$, $(h_v^{(0)})_{i} = 0$ for $i \neq \ell$, $(h_v^{(1)})_{\ell} = 1$ if $G, v \models \varphi_{\ell}$ and $(h_v^{(1)})_{\ell} = 0$ otherwise. For $i \geq 1$, $\mathbf{h}_v^{(i)}$ satisfies the same property.
\end{manualtheorem}

Let $\varphi_{\ell}(x) = \varphi_j (x) \land \varphi_k (x)$, then $\mathbf{C}_{j\ell} = \mathbf{C}_{k\ell} = 1$ and $\mathbf{b}_{\ell} = -1$, then:
\[
(h_v^{(i)})_{\ell} = \sigma \left( (h_v^{(i-1)})_j + (h_v^{(i-1)})_k - 1 \right).
\]
Based on the Definition A.2, we have $(h_v^{(i)})_{\ell} = 1$ if and only if $ (h_v^{(i-1)})_j + (h_v^{(i-1)})_k - 1 \geq 1 $. Then $(h_v^{(i)})_{\ell} = 1$ if and only if
$G, v \models \varphi_j $and $G, v \models  \varphi_{k}$.

Let $\varphi_{\ell}(x) = \neg \varphi_k (x)$, because $\mathbf{C}_{k\ell} = -1$ and $\mathbf{b}_{\ell} = 1$, we have 
\[
(h_v^{(i)})_{\ell} = \sigma \left( - (h_v^{(i-1)})_k + 1 \right).
\]
we have $(h_v^{(i)})_{\ell} = 1$ if and only if $-(h_v^{(i-1)})_k + 1 \geq 1$, which means $(h_v^{(i-1)})_k = 0$. $(h_v^{(i)})_{\ell} = 1$ if and only if $G, v \not\models \varphi_j$, i.e., $G, v \models \varphi_{\ell}$, and $(h_v^{(i)})_{\ell} = 0$ otherwise.

Let $\varphi_{\ell}(x) = \exists^{\geq N} y (r_j(y, x) \land \varphi_k(y))$. Because of $(A_{r_j})_{k\ell} = 1$ and $b_{\ell} = -N + 1$, we have:
\[
(h_v^{(i)})_{\ell} = \sigma \left( \sum_{u \in \mathcal{N}_{r_j}(v)} (h_u^{(i-1)})_k - N + 1 \right).
\]
Let $m = |\{u \mid u \in \mathcal{N}_{r_j}(v) \text{ and } G, u \models \varphi_k\}|$. 
Then we have $(h_v^{(i)})_{\ell} = 1$ if and only if
$\sum_{u \in \mathcal{N}_{r_j}(v)} (h_u^{(i-1)})_k - N + 1 \geq 1$,
which means $m \geq N$. 
Because $G, u \models \varphi_k$ and $u$ is connected to $v$ with relation $r_j$, and $m \geq N$, we conclude that $(h_v^{(i)})_{\ell} = 1$ if and only if $G, v \models \varphi_{\ell}$ and $(h_v^{(i)})_{\ell} = 0$ otherwise.

\end{proof}

\begin{manualtheorem}{Theorem}{2 (Restated)}
	\label{theorem:CGNN2}
A formula $\varphi(x)$ is learned by C-GNN if it can be expressed as a formula in CML.
\end{manualtheorem}
\begin{proof}[proof of Theorem 2]
    \label{proof:cgnn2} 
Assume for a contradiction that there exists a C-GNN that can learn \(\varphi(x)\). Since \(\varphi(x)\) is not equivalent to any formula in CML, according to Theorem C.7 in \cite{understanding} and Theorem 2.2 in \cite{otto2023graded}, there exists two KGs \(G\) and \(G'\) and two entities \(v\) in \(G\) and \(v'\) in \(G'\) such that \(\mathrm{Unr}^L_G(v) \cong \mathrm{Unr}^L_{G'}(v')\) for every \(L \in \mathbb{N}\) and such that \(G, v \models \varphi\) and \(G', v' \not\models \varphi\), where $\mathrm{Unr}^L_G(v)$ represents unraveling tree, for a detailed definition refer to \cite{understanding, ATheory}. Furthermore, according to the definition of RWL-test in \cite{Barcelo00O22}, because \(\mathrm{Unr}^L_G(v) \cong \mathrm{Unr}^L_{G'}(v')\) for every \(L \in \mathbb{N}\), we have \(e_v^{(L)} = e_{v'}^{(L)}\). So, if a formula $\varphi(x)$ is not equivalent to any formula in CML, there is no C-GNN can learn $\varphi(x)$.

\end{proof}

Building on Theorems 1 and 2, we now present a corollary that highlights a specific limitation of C-GNN’s expressiveness.

\begin{manualtheorem}{Corollary}{3 (Restated)}
	C-GNN cannot learn the logical rule structure $U$.
\end{manualtheorem}

\begin{proof}[proof of Corollary 3]

\label{proof:cantU}
    Let $u$ be the source node. At initialization, according to C-GNN, we can learn the unary predicates of CML as $\varphi_{1}(x) = P_{h}(x)$, where $P_{h}$ indicates the initialization label used to distinguish the source node from other nodes in the KG. Then, according to $\varphi_{l}(x)=\exists^{N \geq n}y(\varphi_{k}(y) \wedge r(y,x))$ in CML, we can get $\varphi_{2}(x) = \exists y,\varphi_{1}(y) \wedge r_{1}(y,x)$. At this time, an additional variable $x^{'}$ is needed to indicate that the edges $r_{2}$ and $r_{4}$ come from the same node, that is, 
    \[\varphi_{3}(x,x^{'}) = \exists y,(\varphi_{2}(y) \wedge r_{2}(y,x)) \wedge (\varphi_{2}(y) \wedge r_{3}(y,x^{'})).\] By definition, CML is a first-order logic expression constrained by variables, so $\varphi_{3}$ cannot be expressed in CML, so C-GNN cannot learn the logical rule structure $U$. 

    In addition, if no additional variable $x^{'}$ is added, it can be expressed by two chain logic structures \[C_{3}(x) = \exists y,y_{1},y_{2},\varphi_{1}(y) \wedge r_{1}(y,y_{1}) \wedge r_{2}(y_{1},y_{2}) \wedge r_{3}(y_{2},x) ,\] \[C_{3}^{'}(x) = \exists y,y_{1},y_{2},\varphi_{1}(y) \wedge r_{1}(y,y_{1}) \wedge r_{4}(y_{1},y_{2}) \wedge r_{5}(y_{2},x).\] At this time, $U(x) = C_{3}(x) \wedge C_{3}^{'}(x)$.
According to the definition of CML, the chain logic structure can obviously be expressed in CML. At this time, the logical rule structure $T$ shown in Figure 3 has the same CML expression as $U$. That is, when $\mathcal{G} ,v \models U$ and $\mathcal{G} ,v \models T$, $\mathbf{h}_{v}^{(L)} = 1$ both hold. C-GNN cannot distinguish between $U$ and $T$ under CML, and C-GNN cannot learn the logical expression $U$.	
\end{proof}

\subsubsection{A.2 The Logical Expressiveness of PN-GNN}

This section presents the formal proofs related to the logical expressiveness of PN-GNN. We begin by establishing two foundational lemmas: Lemma 7 shows that PN-GNN can replicate any CML formula learnable by C-GNN, while Lemma 8 demonstrates its ability to capture a more complex logical rule structure beyond C-GNN’s capacity. These lemmas support Theorem 9, which states that PN-GNN is strictly more expressive than C-GNN in terms of its ability to capture logical formulas.


\begin{manualtheorem}{Lemma}{7 (Restated)}
If $\varphi(x)$ is a CML formula that can be learned by C-GNN, then PN-GNN can also learn $\varphi(x)$.
\end{manualtheorem}
\begin{proof}[proof of Lemma 7]
The iterative calculation of PN-GNN can be simplified to the following formula:
\[
\mathbf{h}_{v}^{(t)} = \sigma(\mathbf{h}_{v}^{(t-1)}\mathbf{C} + \sum_{j=1}^{r}\sum_{u \in v}\mathbf{h}_{u}^{(t-1)}\mathbf{A}_{r} + \mathbf{b})
\]
\[
\mathbf{h}_{v} = \sigma(\mathbf{h}_{v}^{(L)} + \sum_{i+j = |d|}(\mathbf{h}_{ij}^{(L)}\mathbf{A}_{pn} + \mathbf{b}_{pn}))
\]

According to Theorem 1
, let $\textbf{A}_{pn}=0$, $\textbf{b}_{pn}=0$ . In this case, the PN-GNN update equation degenerates to the same form as C-GNN, implying that PN-GNN can learn all CML logical formulas. In other words, PN-GNN can learn any formula that is learnable in CML.
\end{proof}

\begin{manualtheorem}{Lemma}{8 (Restated)}
PN-GNN can learn the logical rule structure $U$.  
\end{manualtheorem}

\begin{proof}[proof of Lemma 8]
 Let \[C_{3}(x) = \exists y,y_{1},y_{2},\varphi_{1}(y) \wedge r_{1}(y,y_{1}) \wedge r_{2}(y_{1},y_{2}) \wedge r_{3}(y_{2},x) ,\] \[C_{3}^{'}(x) = \exists y,y_{1},y_{2},\varphi_{1}(y) \wedge r_{1}(y,y_{1}) \wedge r_{4}(y_{1},y_{2}) \wedge r_{5}(y_{2},x) \] be two logic rules, and define $\varphi_{v}(x) = C_{3}(x) \wedge C_{3}^{'}(x)$. If $G,v \models \varphi_{v}$, then $\textbf{h}_{v}^{(L)} = 1$. Next, set $(\mathbf{A}_{pn})_{12}= -1$, $(\mathbf{b}_{pn})_{12} = 1$ in formula (15), and the other values in $\mathbf{A}_{pn}$ and $\mathbf{b}_{pn}$ be 0. Substituting into formula (15), we get \[\mathbf{h}_{v}=\sigma(\mathbf{h}_{v}^{(L)} - \mathbf{h}_{12}^{(L)} + 1).\]
According to formula (9), $\mathbf{h}_{ij}$  is formed by aggregating the representations of the nodes $\mathbf{h}_{w}^{(L)}$ in the path that satisfy the hop distance ($i+j=|d|$) between the head and tail entities. 
Since the node represented by $\mathbf{h}_{w}^{(L)}$ is in $\varphi_{v}$, when $\mathbf{h}_{v}^{(L)} = 1$, according to the induction hypothesis, $\mathbf{h}_{w}^{(L)}=1$. Therefore, $\mathbf{h}_{v} = \sigma(2 - \mathbf{h}_{12}^{(L)})$. 
If and only if there is only one neighbor node with 1-hop from the head and 2-hop from the tail, $\mathbf{h}_{12} = 1$, $(2 - \mathbf{h}_{12}^{(L)}) \geq 1$, so $\mathbf{h}_{v} = 1$. 
Therefore, PN-GNN can learn the $U$ logical rule structure. The logical rule structure  $T$ contains two neighborhoods that are 1-hop away from the head and 2-hop away from the tail, $\mathbf{h}_{w}^{(L)} = 2$, so $\mathbf{h}_{v}=0$ for $T$, that is, PN-GNN can distinguish between the logical rule structure $U$ and $T$.
\end{proof}


\begin{manualtheorem}{Theorem }{9 (Restated)}
\label{theorem:PN-GNN-2}
PN-GNN has strictly more powerful logical expressiveness than C-GNN.
\end{manualtheorem}
\begin{proof}[proof of Theorem 9]
    From theorem 2, we know that C-GNN can learn the logical rule structure if and only if the logical rule structure is in CML. From lemma 7, we can get that the logical expression ability of PN-GNN is not weaker than that of C-GNN. From Corollary 3 and lemma 8, we know that PN-GNN can learn the logical rule structure that C-GNN cannot learn, which is proved.
\end{proof}




To further analyze how the expressiveness of PN-GNN evolves with the number of message-passing hops, we establish two supporting lemmas(10 and 11) that characterize the relationship between $k$-hop and $(k+1)$-hop expressiveness. Together, these results reveal the increasing logical capacity of PN-GNN with deeper propagation and serve as the foundation for Theorem 12.

\begin{manualtheorem}{Lemma}{10 (Restated)}
(k+1)-hop can represent all logical expressions that k-hop can learn.
\end{manualtheorem}
\begin{proof}[proof of Lemma 10]
The expression of $k$-hop is:
\[
\textbf{h}_{v|u,q} = 
{\bigoplus_{{i+j} \leqslant k} 
\text{MLP}_{ij}
(\textbf{h}_{ij})
}
\odot \textbf{h}_{v|u,q}^{(L)}
\]
Here, $ w \in P_{uv}, d(u,w)=i, d(w,v)=j$. 
The expression of $(k+1)$-hop is:
\[
\begin{aligned}  
\textbf{h}_{v|u,q} = 
&{\bigoplus_{{i+j} \leqslant k} 
\text{MLP}_{ij}
(\textbf{h}_{ij})
} \odot \text{MLP}_{(i+1)j}
(\textbf{h}_{(i+1)j})\\
&\odot \text{MLP}_{i(j+1)}
(\textbf{h}_{i(j+1)})
\odot \textbf{h}_{v|u,q}^{(L)}.
\end{aligned}
\]
Assume that the expression is true for $k$-hop. Based on this, let $\text{MLP}_{(i+1)j}(\cdot)$ and $\text{MLP}_{i(j+1)}(\cdot)$ have $ \textbf{A}_{pn}=0$, $\textbf{b}_{pn}=0$ in formula (15), then $k+1$ can also represent the logic, that is, $(k+1)$-hop can learn the logical expression of $k$-hop.

\end{proof}
\begin{manualtheorem}{Lemma}{11 (Restated)}
    (k+1)-hop can learn logical rules that cannot be expressed by k-hop.
\end{manualtheorem}
\begin{proof}[proof of Lemma 11]
According to Lemma 10, $(k+1)$-hop can represent all logical expressions that $k$-hop can learn. Therefore, it suffices to identify a case that is learnable by a $(k+1)$-hop but not by a $k$-hop.
Let \[C_{k+1}(x) = \exists \vec{y},\varphi_{1}(y) \wedge r_{1}(y,y_{1}) \wedge \cdots \wedge r_{k+1}(y_{k},x) \] and \[C_{k+1}^{'}(x) = \exists \vec{y},\varphi_{1}(y) \wedge r_{1}(y,y_{1}) \wedge \cdots\wedge r_{2k+1}(y_{k},x) \] 
be two logic rules, where $\vec{y} = y,y_{1},\dotsc, y_{k}$.
Then we define \[\varphi_{v}(x)=C_{k+1}(x) \wedge C_{k+1}^{'}(x).\] If $G,v \models \varphi_{v}$, the $\textbf{h}_{v}^{(L)} = 1$. Then, we set the $(k+1)$-hop to aggregate the node representations. $(\mathbf{A}_{pn})_{1k}= -1$, $(\mathbf{b}_{pn})_{1k} = 1$ in formula (15), and the other values in $\mathbf{A}_{pn}$ and $\mathbf{b}_{pn}$ be 0. we get \[\mathbf{h}_{v}=\sigma(\mathbf{h}_{v}^{(L)} - \mathbf{h}_{1k}^{(L)} + 1).\] As described in Lemma 8, the node representation $\mathbf{h}_{v}^{(L)}$ is in $\varphi_{v}$, if and only if $\mathbf{h}_{1k} = 1$, $\mathbf{h}_{v} = 1$. Therefore, $(k+1)$-hop PN-GNN can learn the $U$ logical rule structure.

If the number of hops for node aggregation in PN-GNN is 
$k$, as illustrated in Figure A.1, then the representation $\mathbf{h}_{v}$ fails to capture the information of nodes reachable from 
$u$ via the relation $r_1$. As a result, both The logic structure $U$ and the logic structure $T$ yield the same $\mathbf{h}_{v}$, making them indistinguishable to the model. Therefore, $(k+1)$-hop can learn logical rules that cannot be expressed by $k$-hop .

\end{proof}

With Lemmas 10 and 11 established, we are now in a position to prove the main result regarding the logical expressiveness of PN-GNN at different hop levels. Theorem 12 formalizes how this capacity increases with hop depth.


\begin{manualtheorem}{Theorem}{12 (Restated)} 
For PN-GNN, the logical expressive power of (k+1)-hop is strictly more powerful than that of k-hop.
\end{manualtheorem}
\begin{proof}[proof of Theorem 12]
From Lemma 10, we know that $(k+1)$-hop can represent all the logical expressions that $k$-hop can express. In other words, the logical expressiveness of $k$-hop is a subset of that of $(k+1)$-hop.
From Lemma 11, we know that $(k+1)$-hop can represent logical rules that $k$-hop cannot express. This means that $(k+1)$-hop introduces additional expressive power that $k$-hop lacks.
Therefore, the logical expressive power of $(k+1)$-hop strictly exceeds that of $k$-hop, since it combines both the expressiveness of $k$-hop and the new capabilities provided by the extra hop.
\end{proof}

\begin{figure}[ht]
    \centering
  
    \subfigure[(k+1)-hop $T$ structure]{
        \includegraphics[width=0.30\textwidth]{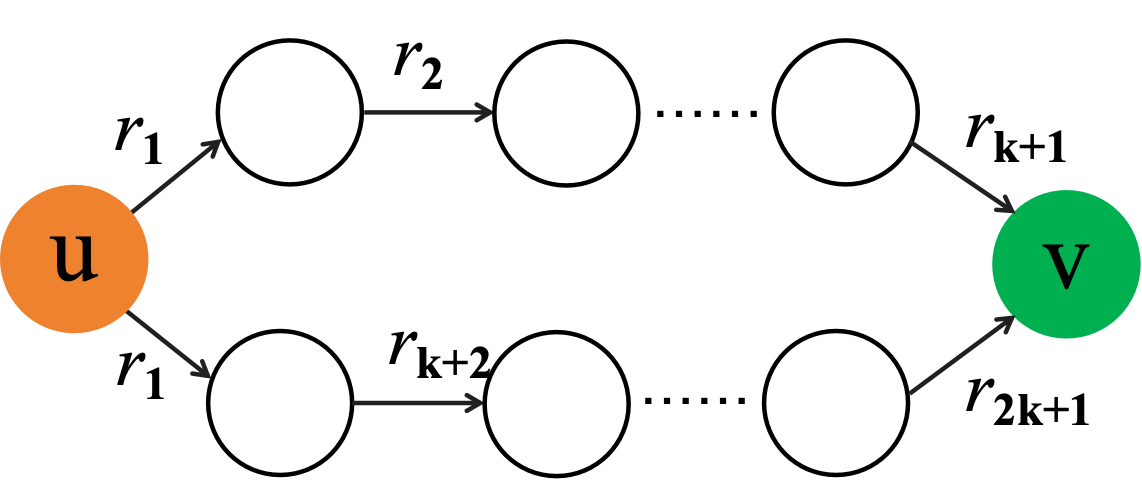}
        \label{fig:aa}
    }
   
    \subfigure[(k+1)-hop $U$ structure]{
        \includegraphics[width=0.30\textwidth]{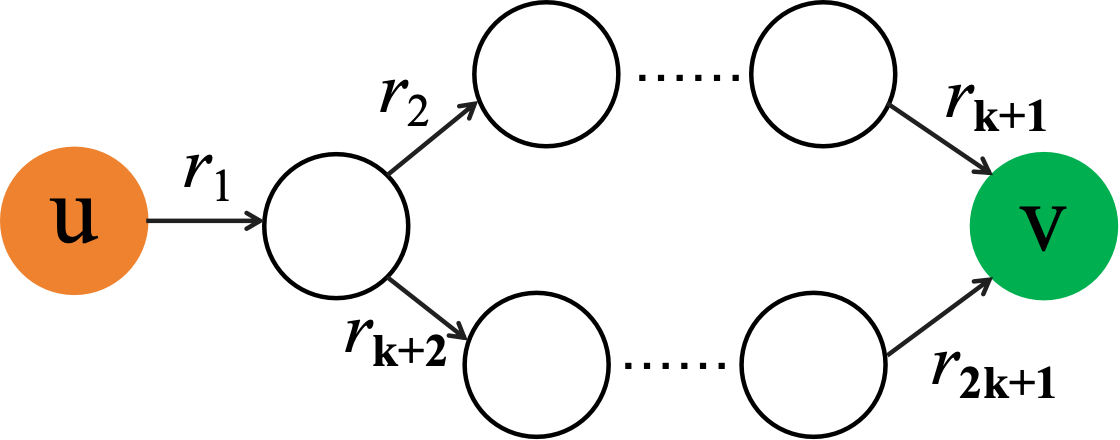}
        \label{fig:bb}
    }
    \caption*{Figure A.1: Schematic diagram of (k+1)-hop $T$ structure and $U$ structure}
    \label{fig:appendix}
\end{figure}
\subsection{B. More details about the experiment}

\subsubsection{B.1 Datasets}
\label{appendix:data}

To evaluate the logical expressive power of PN-GNN, we conducted experiments on eight synthetic datasets and two real-world datasets. 

\subsubsection{Synthetic datasets}

The synthetic datasets $C_{3}$, $C_{4}$, $I_{1}$, $I_{2}$, $T$, and $U$ are adopted from \cite{understanding}, while we additionally construct $T_{label}$ and $U_{label}$ to study the effect of labeling strategies.
The process of generating $T_{label}$ and $U_{label}$ consists of three steps, namely, generation of the logical rule structure, generation of noisy triplets, and completion of missing triplets. 

\begin{itemize}[label= ]

\item 
(1) Following the steps described in \cite{understanding}, we construct the training set and the validation set.
\label{item:1}
\item 
(2) To construct the rule structures for the test set, we introduced additional entities to ensure that none of the entities in these logical structures were assigned constant labels in the training or validation sets.
\label{item:2}
\item 
(3) Missing triplet completion: During the process of adding noisy triplets, some of them may unintentionally align with the target rule structure. These triplets are then included to ensure the completeness of the dataset.
\label{item:3}
  
\end{itemize}
 
These synthetic datasets divide the triplets containing the target relation into training, valid and test. It is particularly noteworthy that all test models only use the known triplets generated in steps \ref{item:1} and \ref{item:2} above, that is, they do not contain the target relation (query relation), thereby preventing the model from making predictions based on the target relation instead of the logical rule structure. 

The statistical information of the synthetic datasets are shown in Table B.1.
\begin{table*}[h]
	\centering
		
		\begin{tabular}{ccccccccc}
			\toprule
            \textbf{Dataset} & $C_{3}$ &  $C_{4}$ & $I_{1}$ & $I_{2}$ & $T$&  $U$ & $T_{label}$& $U_{label}$\\
			\midrule 
			Known Triples & 1514&2013&843&1546&2242&2840&2812&3312 \\
			Train & 1358&2265&304&674&83&396&83&396\\
			Valid & 86&143&20&43&6&26&6&26 \\
			Test & 254&424&57&126&15&183&50&74 \\
			\bottomrule 
		\end{tabular}
        \caption*{Table B.1: The statistical information of the synthetic datasets.}
		\label{sec3:construct_data}
\end{table*}
\subsubsection{Datasets for Transductive Setting}
Transductive setting means that the test set does not contain the target triples in the training set, but all entities and relations are allowed to appear in the training set. We selected two widely used datasets WN18RR \cite{dettmers2018convolutional} and FB15K237 \cite{toutanova2015representing} to verify the effectiveness of the model. The statistical information of the datasets is shown in Table B.2.
\begin{table*}[h]
	\centering

		\begin{tabular}{cccccc}
			\toprule
			\textbf{Dataset} & entities & relations & train & valid & test\\
			\midrule
			FB15K237 & 14541&237&272115&17535&20466 \\
			WN18RR & 40943&11&86835&3034&3134 \\
 	\bottomrule 
		\end{tabular}
        	\caption*{Table B.2: Statistics of transductive benchmark datasets.}
		\label{sec3:real_data}

\end{table*}

\subsubsection{Datasets for Inductive Setting}
\begin{table*}[h]
	\centering
		
		\begin{tabular}{cccccccc}
			\toprule
			& & \multicolumn{3}{c}{\bf{WN18RR}} &\multicolumn{3}{c}{\bf{FB15K-237}}\\	
			&  & relations & entities & links & relations &  entities & links\\
			\noalign{\smallskip}\hline\noalign{\smallskip}
			\multirow{2}{*}{v1}
			& Train  & 9 & 2746 & 6678 & 183 & 2000 & 5226 \\
			& Test & 9 & 922 & 1991 & 146 & 1500 & 2404  \\
			\noalign{\smallskip}\hline\noalign{\smallskip}
			\multirow{2}{*}{v2}
			& Train & 10 & 6954 & 18968 & 203 & 3000 & 12085 \\
			& Test & 10 & 2923 & 4863 & 176 & 2000 & 5092  \\
			\noalign{\smallskip}\hline\noalign{\smallskip} 
			\multirow{2}{*}{v3}
			& Train & 11 & 12078 & 32150 & 218 & 4000 & 22394\\
			& Test & 11 & 5084 & 7470 & 187 & 3000 & 9137 \\
			\noalign{\smallskip}\hline\noalign{\smallskip}
			\multirow{2}{*}{v4}
			& Train & 9  & 3861 & 9842 & 222 & 5000 & 33916\\
			& Test & 9 & 7208 & 15157 & 204 & 3500 & 14554 \\
			\bottomrule
		\end{tabular}
        \caption*{Table B.3: Statistics of inductive benchmark datasets.}
		\label{tab:inductive_statistics}
\end{table*}
Inductive setting refers to the entities in the training set and the test set that do not overlap, but all relations in the test set have appeared in the training set. The widely used WN18RR (v1-v4) and FB15K237 (v1-v4) datasets in the inductive setting were constructed by Komal et al. \cite{teru2020inductive}. The statistics are shown in Table B.3.
\subsubsection{B.2 Hyperparameter Settings}
\label{appendix:hyperparameter}
In the experimental setup in synthetic datasets, the hyperparameter search space is defined as follows: the learning rate is selected from \([0.0001, 0.001, 0.01, 0.1]\), embedding dimensions from \([32, 64, 128, 256, 512]\), number of GNN layers from \([4, 5, 6, 7]\), batch size from \([20, 40, 60]\) and dropout rate within the range \((0, 0.6)\). The activation function is chosen from (\textit{ReLU}, \textit{Tanh}, \textit{IDD}, \textit{Sigmoid}), the aggregation function from \(\{\text{\textit{sum, min, max}}\}\), and the score function adopts either TranE \cite{bordes2013translating} or DistMult \cite{yang2014embedding}.  

For C-GNN and PN-GNN, we employ Ray \cite{liaw2018tune} to perform automatic hyperparameter tuning on the valid. PN-GNN integrates path-neighbors via concatenation to enhance node representations, while C-GNN utilizes \textit{sum} aggregation.

In the transductive setting, the experimental configurations for FB15K237 are as follows: the embedding dimension is set to 64, the number of GNN layers is 6, the relational message passing function adopts \textit{DistMult}, and the aggregation function uses PNA \cite{corso2020principal}. The number of negative samples is 128, meaning each positive sample corresponds to 128 negative samples. The optimizer is Adam \cite{kingma2014adam}, with a learning rate of 0.005 and a batch size of 32. Additionally, we apply LayerNorm \cite{ba2016layer} during training to stabilize the process. For the WN18RR, the embedding dimension remains 64, the number of GNN layers is 6, and the relational message passing function employs \textit{RotatE} \cite{sun2019rotate}. The batch size is set to 16, and the aggregation function remains PNA. The number of negative samples is 32, with the optimizer and learning rate settings kept consistent with FB15K237.  

In the inductive setting, the v1-v4 versions of WN18RR and FB15K237 all employ the Adam optimizer with a uniform learning rate of 0.005. The aggregation function is consistently PNA, the number of negative samples is set to 64, and the batch size is fixed at 64. The embedding dimension is searched within \([32, 64]\), while the relational transfer function is selected from \(\textit{RotatE}, \textit{DistMult}\).  

Across all three dataset settings, PN-GNN utilizes paths with 2-hop and 3-hop, aggregating neighbor representations \(\textbf{h}_{11}\), \(\textbf{h}_{12}\), and \(\textbf{h}_{21}\).  

\subsubsection{B.3 Metrics}
\label{appendix:metrics}
We evaluated the effectiveness of the model using commonly adopted evaluation metrics in reasoning tasks of KG, including mean rank (MR), mean reciprocal rank (MRR), and Hits@k (1,3,10).

\begin{itemize}
    \item \textbf{Mean Reciprocal Rank (MRR)} is an improved version of the Mean Rank (MR) metric. In MR, the model ranks candidate entities (either head or tail) for a given test triple, and the position of the correct entity is recorded as the ranking score. However, MR is sensitive to outliers and may produce unstable results. To address this issue, MRR computes the reciprocal of the ranking score and takes the average over all test queries. Compared to MR, MRR provides better stability and is less affected by extreme values.
    
    \item \textbf{Hits@k} measures whether the correct answer appears in the top-$k$ ranked candidates returned by the model. For example, Hits@1 indicates whether the correct answer is ranked first, while Hits@10 measures if it appears within the top 10 candidates. The results are expressed as percentages (\%).
\end{itemize}

\subsubsection{B.4 Baselines}
\label{appendix:baselines}

The synthetic dataset is to verify the logical expressive power of the KG reasoning method based on GNN. We use two major categories of methods based on R-GNN and C-GNN as comparison baselines. The R-GNN-based methods include R-GCN \cite{schlichtkrull2018modeling} and CompGCN \cite{vashishth2019composition}. The C-GNN-based methods include NBFNet \cite{zhu2021neural} and RED-GNN \cite{zhang2022knowledge}. At the same time, it is also compared with EL-GNN \cite{understanding} based on the labeling strategy.

In the inference setting, we use widely used baselines in the inference setting, namely: representation learning-based, path-based, and GNN-based methods.
\begin{itemize}
	\item Methods based on representation learning:TransE \cite{bordes2013translating}, DistMult \cite{yang2014embedding}, RotatE \cite{sun2019rotate}, HAKE \cite{zhang2020learning}, RotH \cite{chami2020low}, ComplEx+RP \cite{trouillon2016complex} and ConE \cite{bai2021modeling}.
	\item Path-based methods: NeuralLP \cite{yang2017differentiable}, DRUM \cite{sadeghian2019drum}. 
	\item   GNN-based methods: R-GCN \cite{schlichtkrull2018modeling}, CompGCN \cite{vashishth2019composition}, GraIL \cite{teru2020inductive}, RED-GNN \cite{zhang2022knowledge}, A*Net \cite{zhu2024net}, NBFNet \cite{zhu2021neural}, EL-GNN \cite{understanding}.
\end{itemize}

In inductive setting experiments, since representation-based methods need to learn specific entity representations, they cannot cope with the inductive setting. Therefore, path-based methods NeuralLP \cite{yang2017differentiable}, RuleN \cite{meilicke2018fine}, DRUM \cite{sadeghian2019drum}; and GNN-based methods NBFNett \cite{zhu2021neural}, RED-GNN \cite{zhang2022knowledge}, GraIL \cite{teru2020inductive}, CoMPILE \cite{mai2021communicative} are used as comparison baselines. Since EL-GNN needs to label entities with entity representations, it cannot cope with the inductive setting.

\subsection{B.5 Computational Cost}
\begin{table*}[htp]
	\centering	
		\begin{tabular}{ccccc}
			\toprule
			& \multicolumn{2}{c}{\bf{Time(seconds)}} &\multicolumn{2}{c}{\bf{Memory(MB)}}\\	
			& PN-GNN & NBFNet & PN-GNN & NBFNet\\
            \midrule
     v1&23.79&12.00&1130.13&1117.81\\
     v2&	57.00	&30.85&	1825.46&	1803.68\\
     v3&	115.22&	70.64&	2552.14&	2523.39\\
     v4&	197.37&	134.03&	3266.54	&3229.67\\
			\bottomrule
		\end{tabular}
        \caption*{Table B.4: Computational Cost on Different Versions of FB15K237.}
		\label{tab:time-memory}
\end{table*}
We tested the time and memory performance of PN-GNN  and NBFNet on different versions of FB15K237, as shown in Table B.4. The increase in time cost mainly comes from the MLP during aggregation. No extra storage is needed as C-GNN's learned embeddings are used for aggregation. These results show that PN-GNN achieves a favorable trade-off between accuracy and efficiency, especially considering its improved ability to capture complex logical patterns.

\end{document}